\documentclass[twocolum]{raa_twocolumn} 

\usepackage{graphicx,times}
\usepackage{natbib}
\usepackage{amssymb,amsmath}
\usepackage{tabularx}
\usepackage{indentfirst} 
\usepackage{xcolor} 
\usepackage{multirow}
\usepackage[utf8]{inputenc}

\usepackage[pagebackref=true]{hyperref}
\begin{document}

   \title{Sky Background Building of Multi-objective Fiber spectra Based on Mutual Information Network}

 \volnopage{ {\bf 20XX} Vol.\ {\bf X} No. {\bf XX}, 000--000}
   \setcounter{page}{1}

   \author{
        Hui Zhang     \inst{1}, 
        Jianghui Cai    \inst{1,2}, 
        Haifeng Yang       \inst{1,3}, 
        Ali Luo            \inst{4,5},
        Yuqing Yang      \inst{1},
        Xiao Kong       \inst{4,5},
        Zhichao Ding    \inst{6}, 
        Lichan Zhou   \inst{1}, 
        Qin Han      \inst{1}
   }

   \institute{
            School of Computer Science and Technology, Taiyuan University of Science and Technology, Taiyuan 030024, China {\it jianghui@tyust.edu.cn, hfyang@tyust.edu.cn }\\
        \and
             School of Computer Science and Technology, North University of China, Taiyuan 030051, China\\
	\and
            Shanxi Key Laboratory of Big Data Analysis and Parallel Computing, Taiyuan Shanxi 030024, China\\
        \and
            CAS Key Laboratory of Optical Astronomy, National Astronomical Observatories, Beijing 100101, People's Republic of China\\
        \and
            University of Chinese Academy of Sciences, Beijing 100049, People's Republic of China\\
        \and 
            School of science, Liaoning University of Science and Technology, 114051, Anshan, China\\
\vs \no
   {\small Received 20XX Month Day; accepted 20XX Month Day}
}

\abstract{
Sky background subtraction is a critical step in Multi-objective Fiber spectra process. However, current subtraction relies mainly on sky fiber spectra to build Super Sky. These average spectra are lacking in the modeling of the environment surrounding the objects. To address this issue, a sky background estimation model: Sky background building based on Mutual Information (SMI) is proposed. SMI based on mutual information and incremental training approach. It utilizes spectra from all fibers in the plate to estimate the sky background. SMI contains two main networks, the first network applies a wavelength calibration module to extract sky features from spectra, and can effectively solve the feature shift problem according to the corresponding emission position. The second network employs an incremental training approach to maximize mutual information between representations of different spectra to capturing the common component. Then, it minimizes the mutual information between adjoining spectra representations to obtain individual components. This network yields an individual sky background at each location of the object. To verify the effectiveness of the method in this paper, we conducted experiments on the spectra of LAMOST. Results show that SMI can obtain a better object sky background during the observation, especially in the blue end.
}

\authorrunning{H. Zhang et al. }            
\titlerunning{Sky Background Building}  
\maketitle
   
\section{Introduction}\label{sect:intro}

The Large Sky Area Multi-Object Fibre Spectroscopic Telescope (LAMOST) is located at the Xinglong Observatory in Hebei Province, China. It is the first sky survey program in the world to release spectra in excess of 20 million. The availability of large amounts of astronomical data has opened up new opportunities for the application of data science techniques in astronomy. By leveraging the power of big data analytics and machine learning, astronomers can now extract valuable insights from the vast amounts of data generated by multi-target fiber spectroscopy. Such as \citet{33, 34} use machine learning combined with features to classify astrometry object; \citet{41} and \citet{45} use neural networks to detect and classify astrometry object according to the characteristics of astronomical data; \citet{46} analysis the Stellar Spectra from LAMOST by using Generative Spectrum Networks.

Sky background subtraction is a critical step in Multi-objective Fiber spectra process. Large telescopes adopt a similar method for subtracting sky background . Such as LAMOST employs B-spline curve fitting to analyze sky fiber spectra and estimate sky background \citep{38_DR1}. This approach enables the removal of irregularities at specific wavelengths within a given sky fiber during the fitting procedure \citep{1}; Sloan Digital Sky Survey (SDSS) uses a refined model which combines global sky estimation and subsequent localized sky estimation to derive a refined "Super sky" model \citep{4}; Dark Energy Spectroscopic Instrument (DESI) pre-determine fiber flat field correction and resolution matrix elements, iteratively refines these steps to generate a sky spectral model \citep{5}; James Webb Space Telescope (JWST) employs pixel-to-pixel background subtraction and main background subtraction to eliminate extraneous signals and enhance data accuracy \citep{6}. Summarizing from the above methodology leads to that sky information can be effectively acquired through the use of sky fibers.

Recent research on sky subtraction has focused on improving the accuracy of the sky subtracted for telescope characteristics. The methods consider spectrograph stray light, telescope efficacy, fiber transmittance, and wavelength calibration. \citep{14} identified LAMOST stellar spectra sky residuals; \citet{7} found sky estimation initial errors propagate through models, making methods impractical. As spectra processing can introduce flux and wavelength calibration errors, degrading sky subtraction accuracy, \citet{12} and \citet{8} began to analyze 2D data before processing. \citet{10} and \citet{9} proposed methods based on PCA to find better deduction of sky background. \citet{11}, \citet{15} and \citet{16} used non-negative matrix factorization instead of PCA. Furthermore, \citet{13} investigated a method for solving the wide H-[OIII] line in quasars which is affected by the accuracy of the subtraction of the sky; \citet{17} proposed a trend surface-based sky subtraction scheme to address the shortcomings of the bright moonlit night sky model. These methods concentrate only on enhancing the extraction accuracy of sky fibers. Moreover, these methods still fail to take into account the sky information at the target fibers and obtain the precise sky background at a specific location.

As mentioned above, current subtraction mainly relies on sky fibers spectra to build Super Sky. These average spectra are lacking in the modeling of the environment surrounding the objects, so the estimated result is rougher and does not characterize the gradient of the sky background well. The release of large amounts of spectra now also opens up new perspectives for subtracting sky background, allowing us to capture more detailed sky information. This paper first analyzes the sky background components in the target and sky fibers, which include the common information of sky between multiple spectra (the sky continuum spectrum), and the unique information about the spectral location (sky emission lines). Then, we maximize the mutual information between representations derived from different spectra to obtain the common component and minimizes the mutual information between different spectra representations that contain common and exclusive to learn exclusive components. Finally, obtain the full sky background.

This paper is organized as follows: Section \ref{sect 2} introduces a sky background estimation model based on mutual information. Section \ref{sect 3} introduces our dataset, then the results of the study and the analysis are presented. Along with a summary of the research work in this article is provided in Section \ref{sect 4}. 

\section{Method} \label{sect 2}

\begin{figure*}[!ht]
    \centering
    \includegraphics[width=0.8\textwidth]{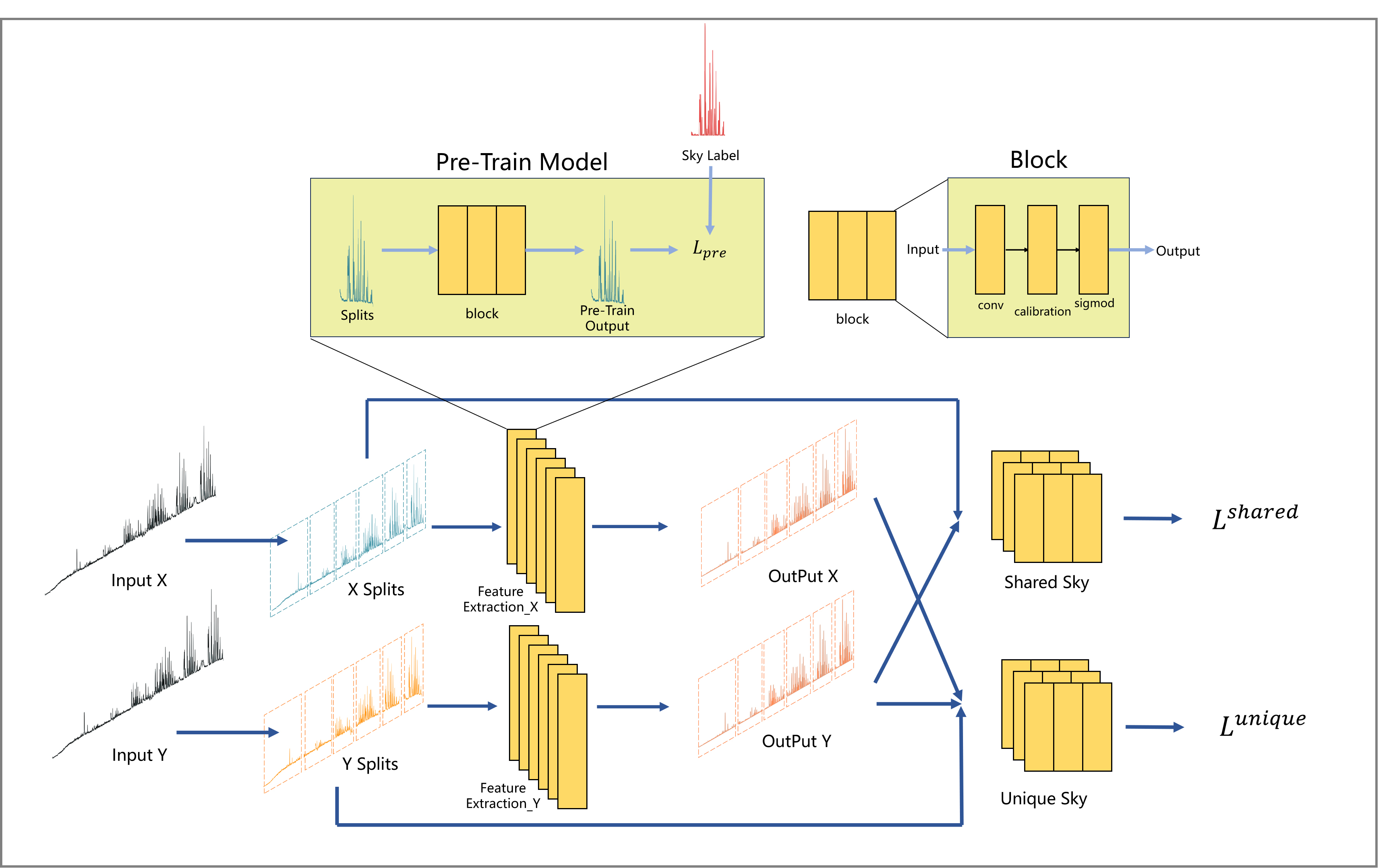}
    \caption{SMI: Sky background construct model based on mutual network. This model includes a block for conducting convolutional operations on spectral data. A pre-trained model is then utilized with sky labels in the training process, involving two stages. During training, the data is adaptively segmented based on the density of sky emission lines to generate segments. In the initial stage, representations are exchanged to calculate mutual information for training and derive a "shared" representation. In the subsequent stage, the objective function is modified for training to obtain a "unique" representation.}    
    \label{FIG:3}
\end{figure*}

In this paper, the proposed model SMI focuses on sky background estimation based on mutual information. The structure of SMI is shown in Figure \ref{FIG:3}. To estimate the sky information more accurate, we first analysis the spectra to two components from a computer point of view (Section \ref{sect 2.1}).  Furthermore, considering the high dimensionality and small feature span characteristics of the spectra, propose a convolution module as an alternative to 1D-convolution for network architecture construction, it contains a calibration module to address the feature alleviating during the estimation network (Section \ref{sect 2.2}). In the two-stage training process, SMI first processes paired spectra data with extracting sky emission line features. These features are then exchanged to compute the mutual information between the spectra data and the sky features. The network is trained by maximizing this estimated MI, yielding a “shared” representation. In the second stage, the mutual information between the input spectra data and the “shared” representation from the first stage is estimated. Simultaneously, the mutual information between sky representations of both spectra data are estimated. A specific sky background is obtained through a linear combination of these two estimates. Additionally, a constraint is applied to minimize discrepancies between sky backgrounds at different spatial locations (Section \ref{sect 2.3}). 

\subsection{Sky Background Spectra Component Analysis} \label{sect 2.1}

The main sources of sky background include moonlight, starlight, zodiacal light, atmospheric scattering and absorption, and the solar activity cycle. Of these, moonlight is an important component, and since computers are affected by different features of moonlight when processing spectra, we analysis moonlight and analyze mainly its features.

\begin{figure}
    \centering
    \includegraphics[width=0.9\columnwidth]{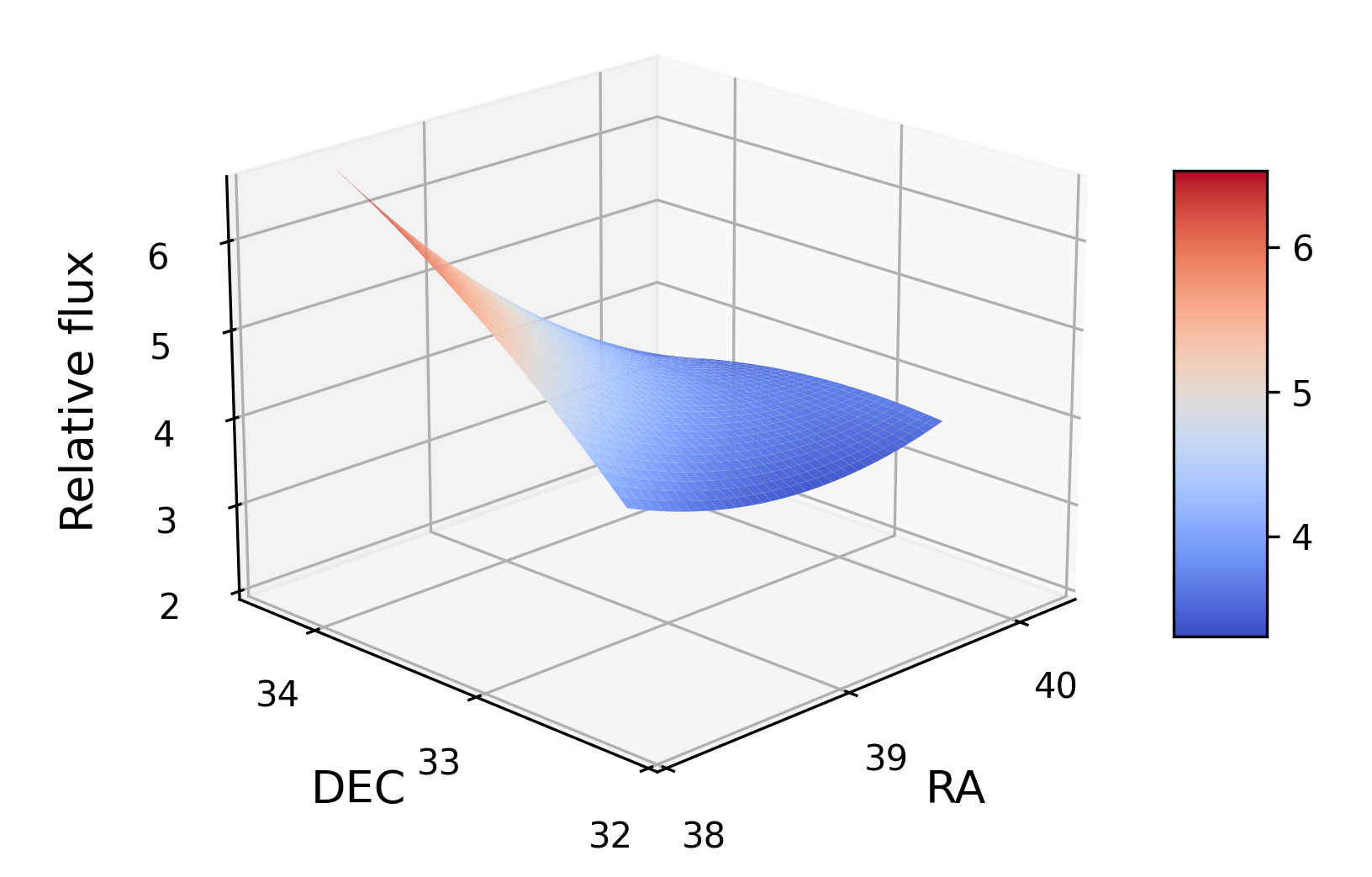} 
    \caption{Sky background trend surface. The spatial position is denoted by (x, y), where x represents the right ascension (RA) coordinate of the fiber in space, and y represents the declination (DEC) coordinate of the fiber in space. The flux indicates the relative flux of the spectrum at the specific declination of the equinox. It is clear that the sky background flux varies slowly over a given region}    
    \label{FIG:1}
\end{figure}

\begin{figure*}[ht]
    \centering
    \includegraphics[width=1.0\textwidth]{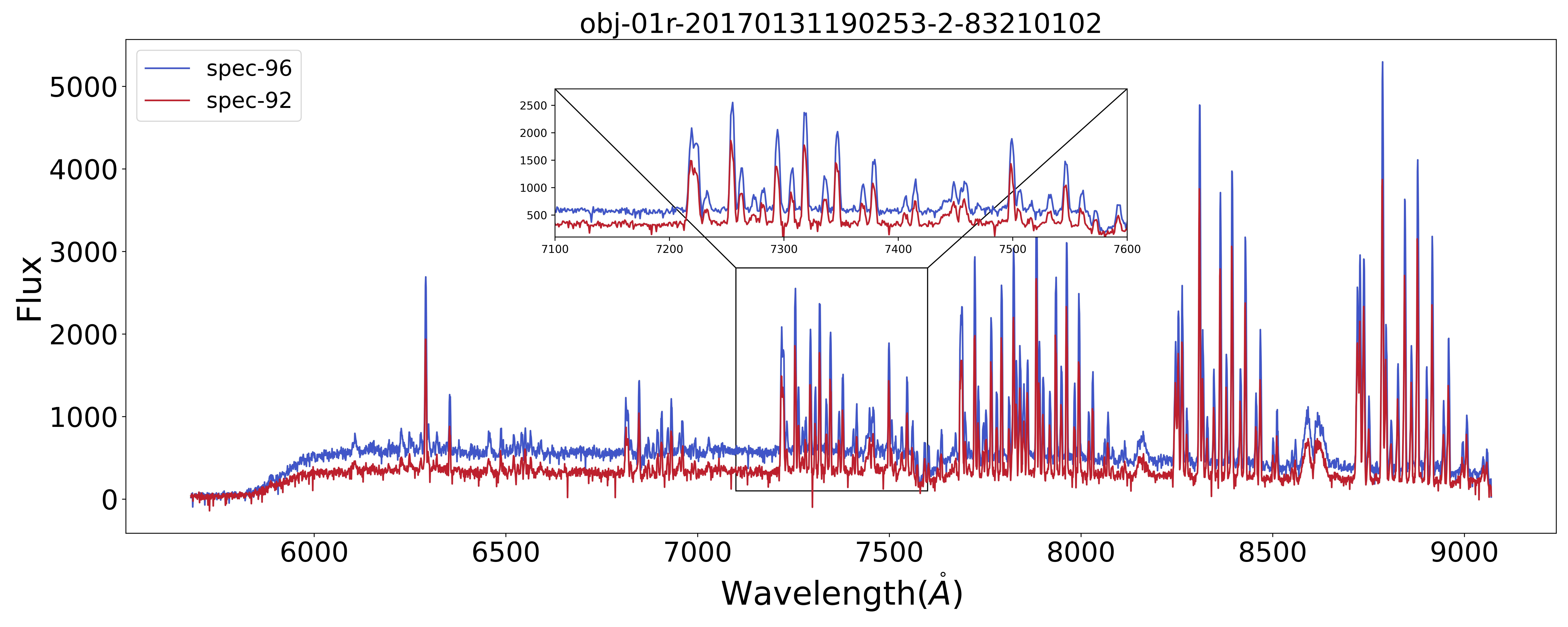} 
    \caption{Spectral data for positional proximity. The distance between spec-96 and spec-92 is 0.5°. An observable distinction in the sky background emission flux, within the spectral range of 7100\AA\ - 7600\AA\, is noted when comparing the data collected by the fiber optics of spec-96 and spec-92.}    
    \label{FIG:2}
\end{figure*}

In a single observation of the multi-target fiber, the sky background within a field is essentially consistent while slightly variable at different positions, as mentioned in \citet{8} and \citet{17}. Specifically, on the spatial scale, moonlight has an overall gradient distribution and can be regarded as a gradually changing surface. As depicted in Figure \ref{FIG:1}, in this area, the relative flux of moonlight is higher on the left side and lower on the right side, with a slow change in the middle. The sky background is analyzed to two components as it consists mainly of a continuum spectrum and emission lines. Within a given sky region, the continuum show little variability, and their errors have a negligible effect on the assessment of target elements. As shown in Figure \ref{FIG:2}, the emission lines show variability, with the position and intensity of sky emission lines varying from location to location. When performing operations like convolution on spectra, the feature extraction of the continuous spectrum is not satisfactory. This also has an impact on the feature extraction of emission lines and subsequent operations. Consequently, we analysis the spectra here. The sky background is decomposed into continuum and emission line components, which are estimated using distinct method. Concurrently, for target optical fibers, it contains both target stellar spectra and sky background, necessitating their explicit separation to isolate astronomical objects of interest. This paper focuses on estimating the emission line components rather than the entire sky background. Thus, the spectra analyzed for a specific region is as follows.

Each sky spectrum is a function of the coordinates and wavelength in the observation sky region. Equation \ref{Eq:4} symbolizes the spectrum in the $i^{th}$ fibe before sky background subtraction step,
\begin{equation}
    I=O_o ( i,\lambda )* d_{sky} ( \lambda )+Sky( \lambda )
\label{Eq:4}
\end{equation}
\noindent where $\lambda$ indicates the wavelength, $O_o (i,\lambda)$ represents the spectral information of the target object captured by the $i^{th}$ fiber, $d_{sky} (\lambda)$ represents the extinction of the Milky Way in space and the atmosphere at a specific wavelength, with minimal variation within a smaller sky region. $Sky(\lambda)$ represents the sky background, including starlight, moonlight, and atmospheric emission lines. We focus on the sky information in the observed data and simplify the spectra as follows:
\begin{equation}
    O( i,\lambda )=[ O_o ( i,\lambda )+S( i,\lambda )  ]  H( i,\lambda )
\label{Eq:5}
\end{equation}

\noindent where $O( i,\lambda )$ is the observed data ignoring the effect of Galactic extinction, and $H( i,\lambda )$ represents the difference between the fibers at the time of the shot. We employ the intensities of sky emission lines at 5577\AA\ to address the difference of fiber efficiency, like it was done in \citet{17}. $S( i,\lambda )$ means the $Sky( \lambda )$ spectrum in the $i^{th}$ fiber without other factor like Milky Way and the atmosphere. The specific equation is described as follows:
\begin{equation}
    S( i,\lambda )= S_m ( i,\lambda )+ S_o ( i,\lambda )
\label{Eq:6}
\end{equation}

\noindent where $S_m ( i,\lambda )$ represents the common sky background within the sky region, including relatively constant continuous spectra and some emission line information; $S_o ( i,\lambda )$ represents the exclusive sky background at the specific position, including the emission line.
\begin{equation}
   S_m  ( i,\lambda )=S_l ( i,\lambda )+S_{sm} ( \lambda )
\label{Eq:7}
\end{equation}

\noindent where $S_l ( i,\lambda )$ represents continuous spectra in the common sky background and $S_{sm} ( \lambda )$ represents the emission line. The whole spectra of this area is used to calculate the $S_l ( i,\lambda )$. The equation for $S_l ( i,\lambda )$ is shown as follows:
\begin{equation}
    S_l ( i,\lambda )=  \frac{1}{N} \sum_{i}^N f(i, \lambda)
\label{Eq:8}
\end{equation}

\noindent $f( i,\lambda )$ represents the continuous spectral information of fiber $i$. It is obtained by averaging continuous spectra in the neighborhood, and $S_{sm} ( \lambda )$ represents the component of sky emission lines in the common sky information.

Then an observed spectrum can be represented as:
\begin{equation}
\begin{aligned}
    O( i,\lambda )=[ O_o ( i,\lambda )+S_l ( i,\lambda ) + S_{sm} ( \lambda ) + 
    \\ S_o ( i,\lambda )]  H( i,\lambda )
\label{Eq:9}
\end{aligned}
\end{equation}

For target fibers, it comprises two components: the desired signal from the target star and contaminating sky background noise. In contrast, sky fibers contain only sky background emission, providing a reference for accurate sky modeling and subsequent subtraction in target fiber data.
\begin{equation}
\begin{aligned}
    O( i,\lambda )=[ S_l ( i,\lambda )+S_{sm} ( \lambda ) + S_o ( i,\lambda ) ]  H( i,\lambda )
\label{Eq:n1}
\end{aligned}
\end{equation}
\noindent where $H( i,\lambda )$ has the same meaning as described in the previous, indicating the differences between different fibers; $S_l ( i,\lambda )$ represents the spectra data continuum spectrum information, which is obtained through the median of the neighborhood; $S_{sm} ( \lambda )$ represents the common sky background within the sky region, including a relatively constant continuum spectrum and emission line information; $ S_o ( i,\lambda )$ represents the exclusive sky background at a specific location, which is mainly embodied in the emission lines.

Based on the above, we can obtain $S_{sm} ( \lambda )$ and $ S_o ( i,\lambda )$ from different fiber data by mutual information estimation, which enables the estimation of the sky background at the target fiber and sky fiber.

\subsection{Feature  Extraction of spectral data} \label{sect 2.2}

SMI focuses on the sky emissions component, so we design a feature extraction module aimed at extracting emission features from the spectra data. The sky feature is presented as an emission line feature with a wavelength width of approximately a dozen units, and the distribution of this feature is uneven. This may lead to the potential loss of sky information at various locations when using multi-layer convolution. Moreover, the positions of the extracted sky emission lines may deviate, which complicates the convergence and descent of the loss function. 

\begin{figure}[!ht]
    \centering
    \includegraphics[width=1.0\columnwidth]{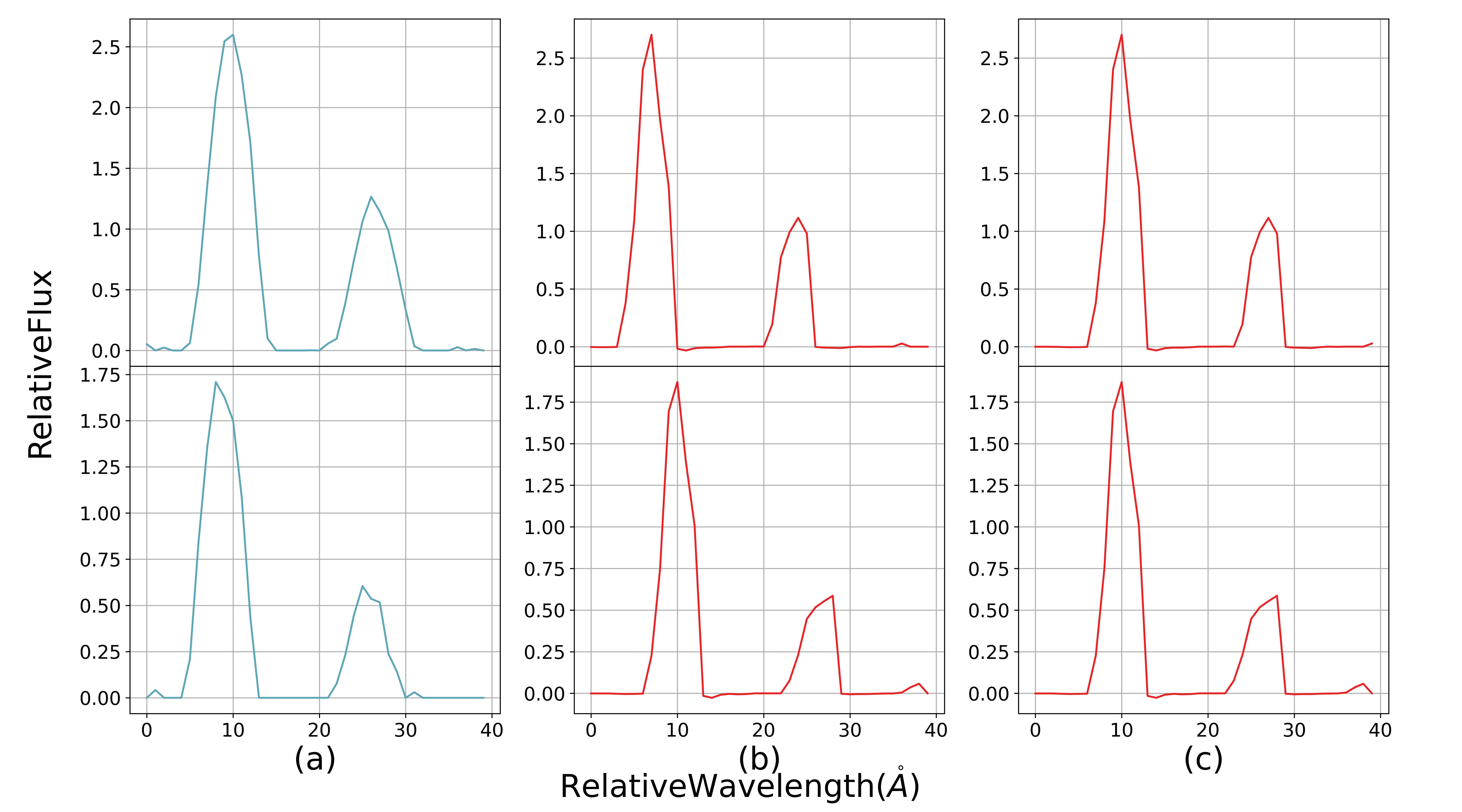} 
    \caption{Deviation of spectral data after convolution. A set of spectral data (a) after convolution operation results in data set (b), which has positional shifts in the emission line features. The data set (c) is obtained after the calibration module.}    
    \label{FIG:block}
\end{figure}

\begin{figure}
    \centering
    \includegraphics[width=1.0\columnwidth]{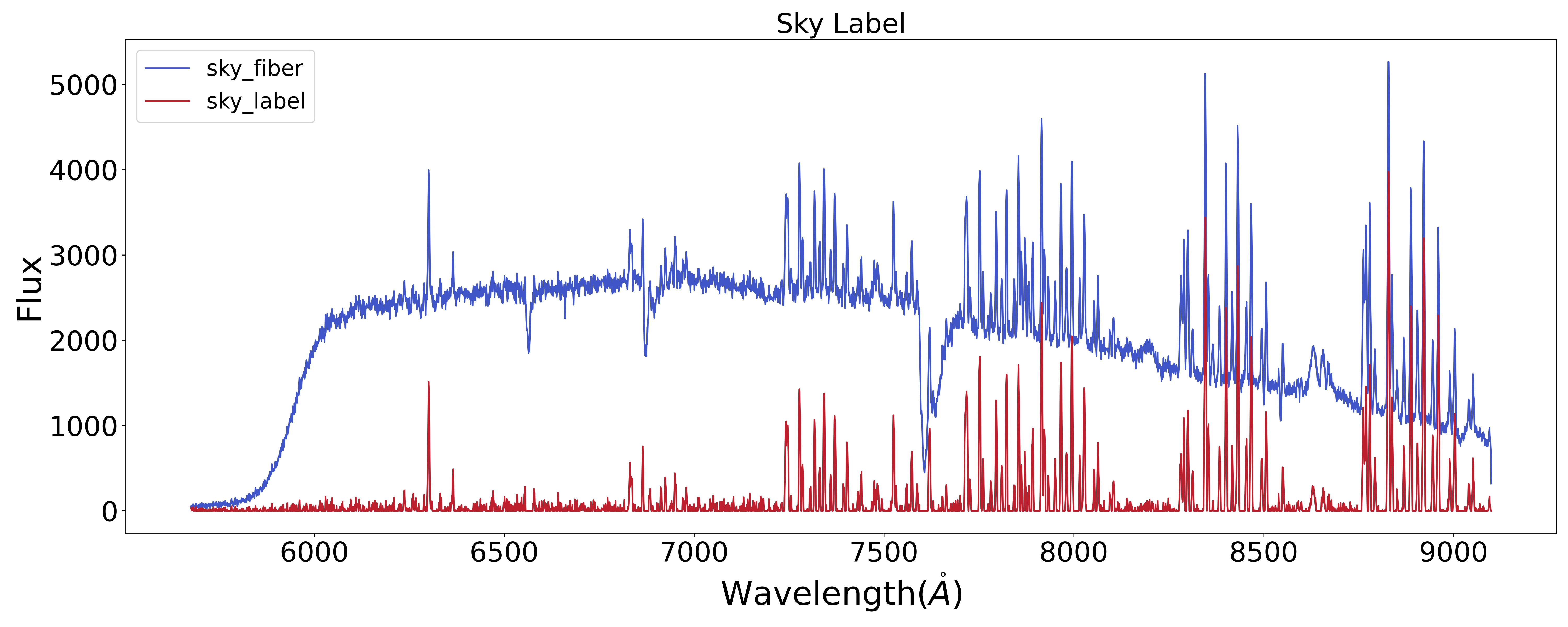}
    \caption{
    Sky Label. The spectral data is depicted by the blue solid line, and the sky label obtained through the neighborhood median method represents as the orange solid line. It can be seen that the continuous spectrum of the observed spectra is removed and only the emission line part is retained.
    }    
    \label{FIG:4}
\end{figure}

To address this issue, this study introduces a block consisting of convolution model, calibration module and activation function. The convolution module is used to extract data features. We use a small convolution kernel to extract spectral feature. Subsequently, calibration model uses anomaly detection to detect the emission lines in the spectral data, aiming to align the features so as to solve the problem of shifted emission line features. As depicted in Figure \ref{FIG:block}, the deviation of emission line features is shown in (b), a comparative of flux and relative positioning reveals correspondence between the first emission line characteristics in (a). However, the peak position at coordinate 10 in bottom (b), while exhibits an offset in upper (b). This has an impact in mutual information estimation due to impaired position. Application of wavelength calibration effectively mitigates this discrepancy, as evidenced in (c). The calibration module, which is applied to the pre-train module and the mutual information calculation module, is designed to accelerate the convergence of the loss function in each part and effectively enhance the training accuracy.

\subsection{Sky Background Estimation Based on Mutual Information} \label{sect 2.3}

\begin{figure}
    \centering
    \includegraphics[width=1.0\columnwidth]{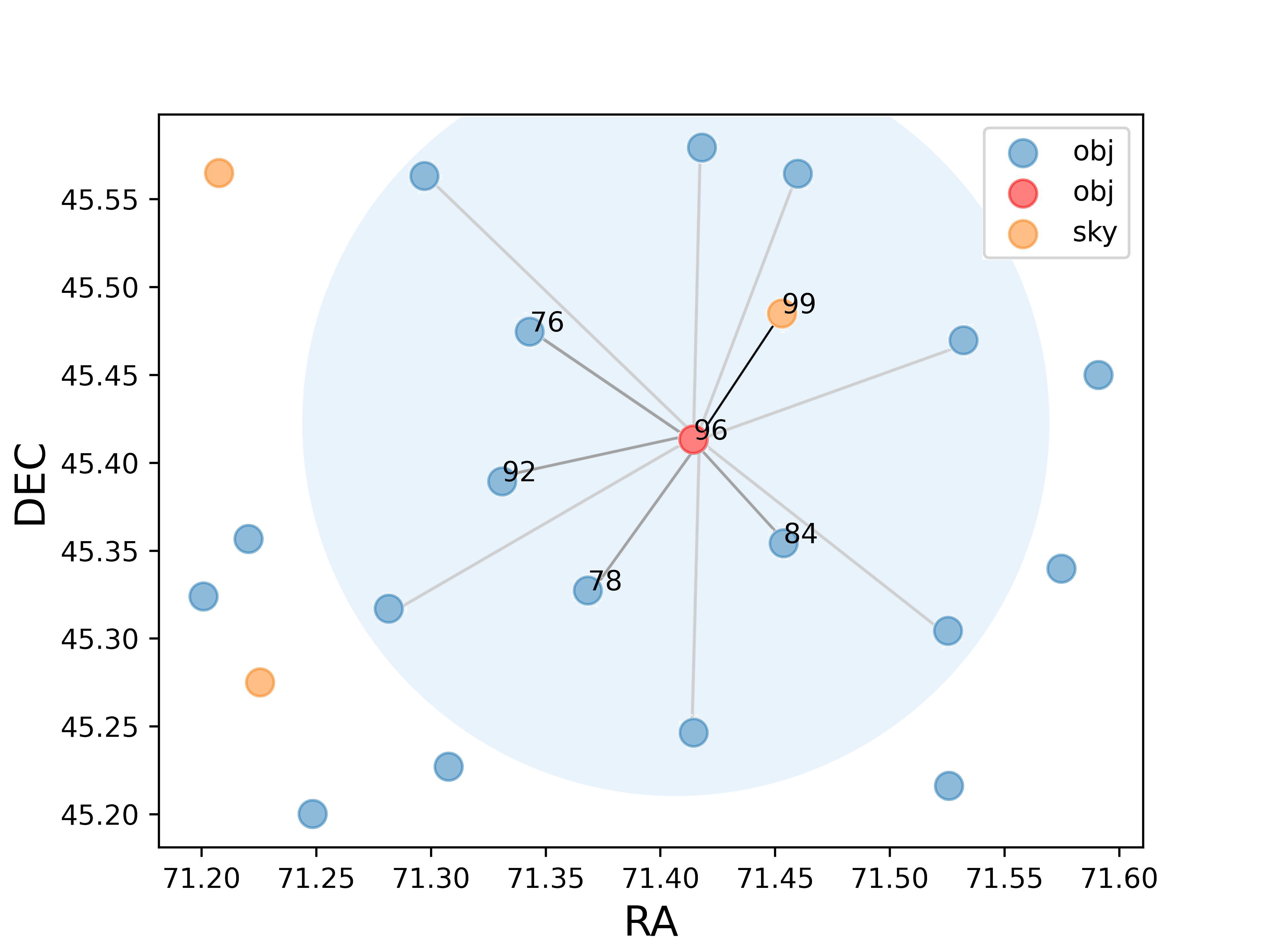}
    \caption{
    Schematic diagram of mutual information-based object sky background construction. Several fiber at plan 20170130GAC071N47B1-01r. Where distinct color denoting different categories. Specifically, blue dots represent target fibers, orange dots represent sky fibers. The depth of the line indicates the amount of information.
    }    
    \label{FIG:diff_dot}
\end{figure}

Figure \ref{FIG:diff_dot} shows the distribution of several fibers in a single observation to show the object sky background construction based on mutual information. Taking fiber 96 as an example, SMI firstly inputs all fiber spectra into the network, then maximize mutual information between representations to capturing the common component, known as shared sky of this region. As it can be seen, closer fiber provides more information. In the second stage, fiber 96 and other fiber are inputted into the network for minimizing the mutual information between representations to obtain individual components. Finally the sky background of object fiber 96 is obtained. As depicted above, sky background can be described as Equation \ref{Eq:9}. Based on this, our model aims at estimate $S_{sm} ( \lambda )$ and $S_o ( i,\lambda )$. As shown in Figure \ref{FIG:3}, the data is divided into segments with unequal length for subsequent training, and each segment corresponds to a pre-train model. The pre-train model and the subsequent model consist of feature extraction mentioned before. For the two different objectives, the model obtains the "shared" sky representation $S_{sm} ( \lambda )$ by maximizing the mutual information of multiple sky features, and then minimizes the mutual information between the "shared" and "unique" representations to achieve the learning of the exclusive component $S_o ( i,\lambda )$.

\subsubsection{Pre-Train Model} \label{sect 2.3.1}

Before obtaining the sky representation through mutual information, it is necessary to first capture the emission lines in the spectrum. Moreover, we need to process the celestial components in the spectral data to avoid having residual celestial components in the obtained sky background, which could otherwise affect the results. Consequently, we trained a pre-trained model. 

The pre-trained model is trained using data from the central six spectrometers among the 16 spectrometers employed in a single observation. Spectral data containing information on sky emission lines were used as sky labels. Firstly, the information about the celestial data at the corresponding positions was removed. Secondly, continuous spectral information was removed by the neighborhood median method as described in a previous study \citep{9}. Figure \ref{FIG:4} illustrates a sky spectra and sky label of the spectrometer. The feature extraction network is structured using a three-layer block, with DeepInfomax serving as the loss function. Let $En^{pre}$ denotes the pre-training model function, the spectral data $X$ is inputted into the model, and the objective function is derived as follows:
\begin{equation}
    L_{pre}=KLDivLOSS ( SKY_{label}  ,En^{pre} (X)  )
\label{Eq:11}
\end{equation}

\subsubsection{Mutual Information} \label{sect 2.3.2}

Mutual information is used to measure the correlation between two variables and is widely used in fields such as data science. It is able to capture both linear and nonlinear relationships. By calculating the mutual information between two features, the degree of correlation can be assessed. Let $X$ denotes the observed spectra and $Z$ denotes the sky representation extracted from another observed spectra (in the second stage $Z$ is denoted as $X$ corresponding to the sky representation), $p(x,z)$ is the joint probability density function of $X$ and $Z$, and $p(x)$ and $p(z)$ be the corresponding marginal probability density functions, then the correlation between $X$ and $Z$ is as follows:
\begin{equation}
    I(X,Z)=\int_{x} \int_{z} p(x,z)log(\frac{p(x,z)}{p(x)(z)})dxdz
\label{Eq:1}
\end{equation}

The theoretical mutual information depicted above relies on probability density functions that are difficult to obtain in real-world. Here, we focus on an estimator called MINE, as it performs well in estimating mutual information for continuous random variables in high-dimensional spaces \citep{22}. MINE leverages the Donsker-Varadhan representation of the Kullback-Leibler scattering (KL scattering), by parameterizing a lower bound of the KL divergence and continuously training to improve the lower bound for estimating mutual information.
\begin{equation}
    D_{KL}(X||Y)=
    \sup \limits_{T \colon \Omega \rightarrow \mathbb{R} }
    \mathbb{E}_\mathbb{X}[T_\theta ] 
    - \log_{}{(\mathbb{E}_\mathbb{Y}[e^{T_\theta}])}
\label{Eq:2}
\end{equation}

\noindent where $T_\theta$ represents the neural network function, and different $T_\theta$ are obtained by optimizing $\theta$ to approach the value of mutual information. Therefore, the expression $I_\Theta (X,Z)$ for information measurement via neural network is given as follows:
\begin{equation} \label{Eq:3}
\begin{aligned}
    I(X,Z) &\geq I_\Theta (X,Z)
    \\&=\sup \limits_{\theta \in \Theta }
    \mathbb{E}_\mathbb{P_{(X,Z)}} [T_\theta] - 
    \log_{}{(\mathbb{E}_{\mathbb{P}_X \times \mathbb{P}_Z} [e^{T_\theta}])}
\end{aligned}
\end{equation}

The equation has been proven to yield stable parameter results, leading to stable mutual information results. A statistical network, similar to that used in Deep InfoMax \citep{47}, is employed to estimate the mutual information value $\theta$.

\subsubsection{Sky Background Estimation} \label{sect 2.3.3}

At the first stage, we estimate the “shared” representation by maximizing the mutual information. Let $En_{\varphi X}^{sh}: X \rightarrow SX$ and $En_{\varphi Y}^{sh}: Y \rightarrow SY$ denote two pre-trained module with the same initial parameters. They extract the sky emission line features $SX$ and $SY$ from the spectral data $X$ and $Y$ respectively, and $Th_{\theta X}$, $Th_{\theta Y}$ denote the mutual information estimators. As suggested in a previous study \citep{31}, a method focuses solely on the common information between the sky features $X$ and $Y$. This involves switching the "shared" representation to calculate the cross-mutual information, as illustrated in Equation \ref{Eq:12}. The process of switching the representation is crucial in deriving the "shared" representation, as it facilitates the removal of unique information pertaining to each sky feature.
\begin{equation}
    L_{MI}^{sh} = Th_{\theta X,\varphi Y} (X,SY) + 
    Th_{\theta Y,\varphi X} (Y,SX)
\label{Eq:12}
\end{equation}

Furthermore, it is necessary for the sky representations to possess identical "shared" representations, thus a constraint is introduced to reduce the $L1$ distance between "shared" representations.
\begin{equation}
    L_1^{sh}=|| SX - SY ||
\label{Eq:13}
\end{equation}

The objective function of the final learning "shared" representation is a linear combination of the previous loss terms, with $\alpha$ is a constant factor.
\begin{equation}
    \max \limits_
    { \left\{\theta,\varphi \right\} X, Y}
    L^{shared}=  L_{MI}^{sh} - \alpha L_1^{sh}
\label{Eq:14}
\end{equation}

After the first stage, the model can obtain "shared" representation $SX_{sm}$ and $SY_{sm}$. Using these model as the initial model for the second stage, the second stage acquires exclusive component $S_o ( i,\lambda )$. Let $En_{\varepsilon X}^{un}: X \rightarrow EX$ and $En_{\varepsilon y}^{un}: Y \rightarrow EY$ be the exclusive information extraction network functions, the representations that includes “shared” and “exclusive” as $EX$ and $EY$ from the spectral data $X$ and $Y$. This focuses on computing the mutual information between the corresponding representations, specifically $X$ and $EX$ , as well as $Y$ and $EY$, without involving any cross-representations.
\begin{equation} \label{Eq:15}
\begin{aligned}
    L_{MI1}^{un} = & Th_{\theta X,\epsilon X} ( X,En_{\varepsilon X}^{un}(X))+ Th_{(\theta Y,\epsilon Y)} (Y ,En_{\varepsilon Y}^{un}(Y))
\end{aligned}
\end{equation}

It is also further restricted by calculating the mutual information between the two features $EX$ and $EY$, aiming to prevent the incorporation of redundant information in the learned representation.
\begin{equation}
    L_{MI2}^{un}= Th_{\theta X,\epsilon X, \epsilon Y}(EX ,EY)
\label{Eq:16}
\end{equation}

The objective function of the final learning “unique” representation is a linear combination of the previous loss terms, as shown in Equation \ref{Eq:17}, where $\beta$ is a constant factor.
\begin{equation}
    \max \limits_
    { \left\{\theta,\epsilon \right\} X, Y}
    L^{unique}=  L_{MI1}^{un} - \beta L_{MI2}^{un}
\label{Eq:17}
\end{equation}

\section{Experiments} \label{sect 3}

In this section, the data preprocessing is introduced. and then analyze the model results from three aspects: Accuracy of the whole spectrum; Accuracy of the sky emission lines and Visualization of "shared" sky background results.

\begin{figure}[!ht]
    \centering
    \includegraphics[width=1.0\columnwidth]{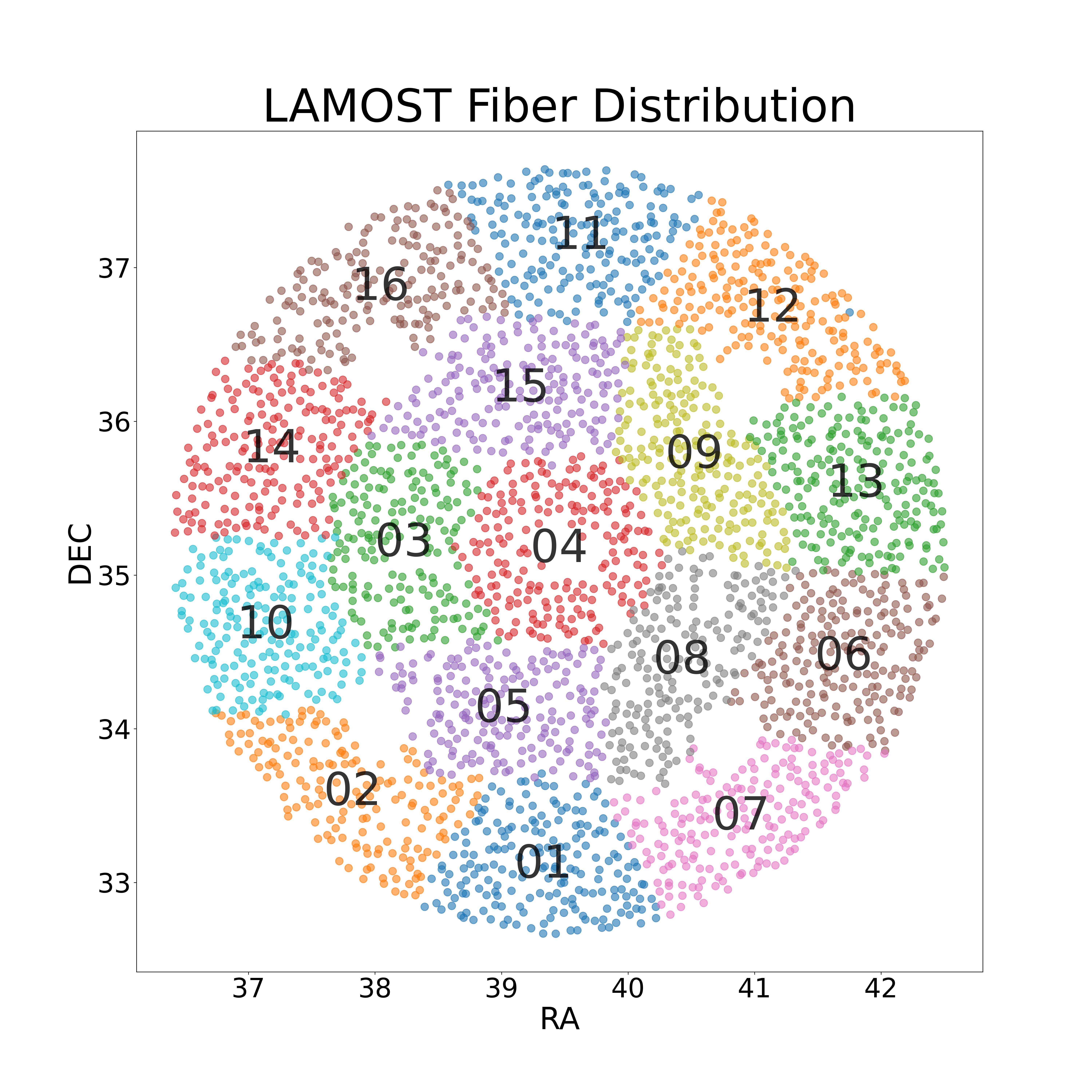}
    \caption{
    Distribution of fiber optic shooting positions in the focal plane. The fiber within the focal plane is divided into 16 regions, with each region linked to a specific spectrometer. The fiber locations are denoted by solid dots, and the corresponding spectrometer identifiers are labeled at their respective positions.
    }    
    \label{FIG:5}
\end{figure}
\vspace{-0.3cm}

\begin{figure}[!ht]
    \centering
    \includegraphics[width=1.0\columnwidth]{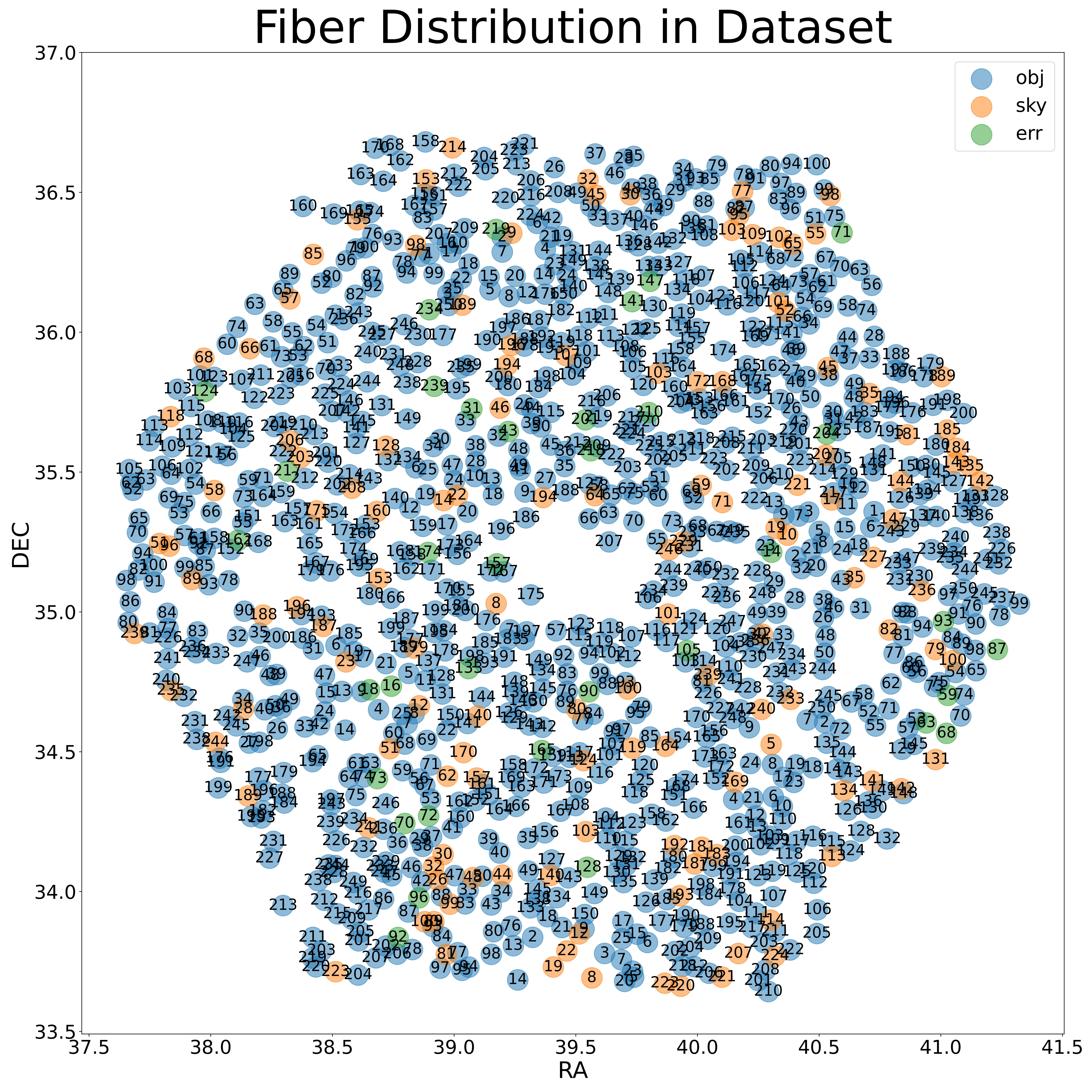}
    \caption{
    Distribution of spectrometer fiber shot locations.
    Where distinct color denoting different categories. Specifically, blue dots represent target fibers, orange dots represent sky fibers comprising approximately 20$\%$ of the total, and green dots represent faulty fibers that are excluded from sky background estimation in this particular observation.
    }    
    \label{FIG:6}
\end{figure}

\begin{figure}[!ht]
    \centering
    \includegraphics[width=1.0\columnwidth]{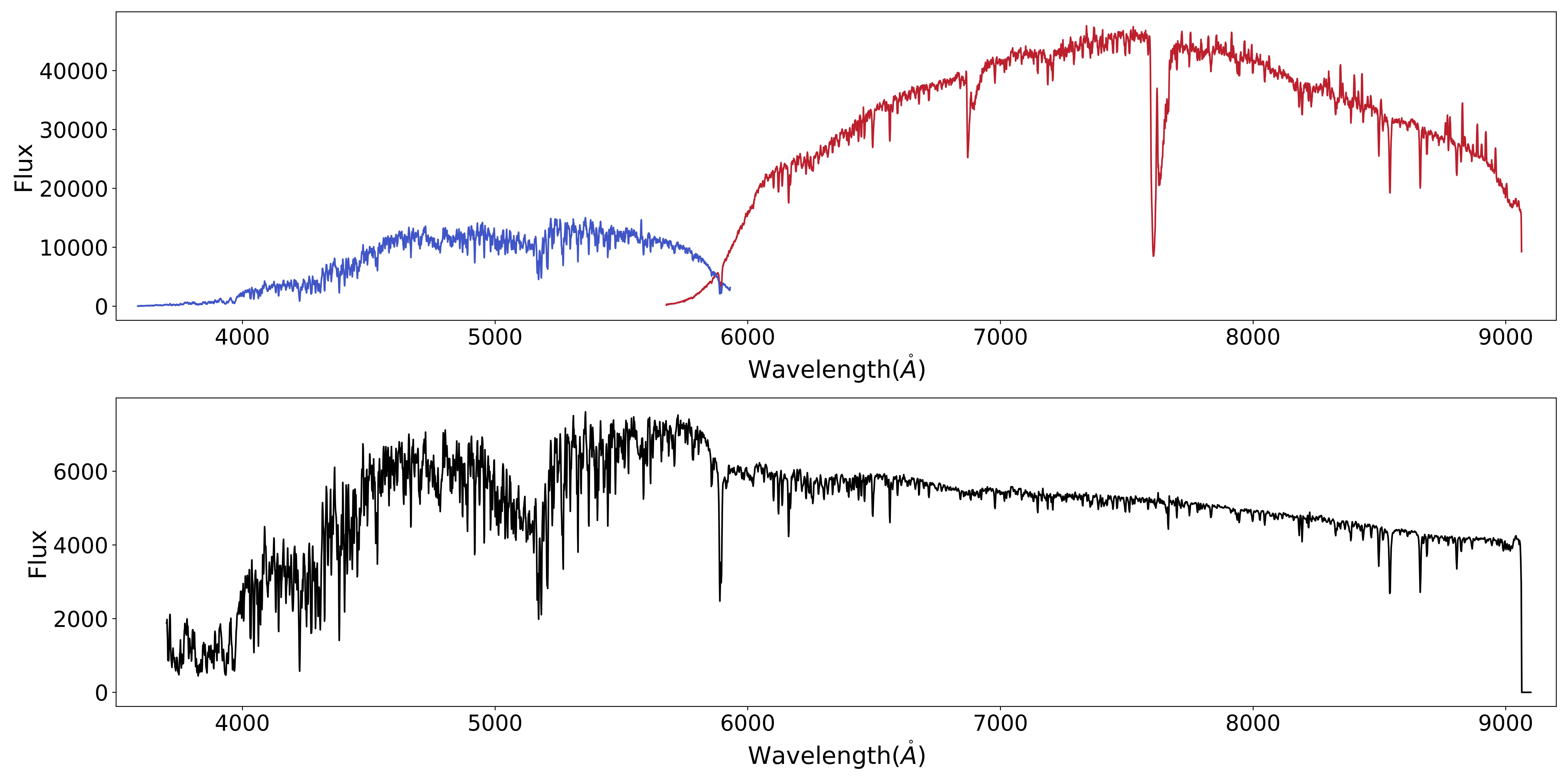}
    \caption{
    The same fiber data between the LAMOST released version and the data from this study. The upper panel displays the data utilized in this study, with red and blue curves representing red-arm and blue-arm spectra data respectively. The lower panel presents the corresponding spectra data officially released by the LAMOST group.
    }    
    \label{FIG:new}
\end{figure}

\subsection{Data Reduction}

Considering that Multi-objective Fiber Optic Telescope has a large field of view, the sky gradient becomes significant. However, using only sky fibers is inadequate to depict the variation of this gradient. Moreover, sky information also exists in the target fibers. Consequently, this model is trained with all fibers in a single observation. Due to the large scale of a single observation, the estimation is initially chosen several spectrograph data. We use LAMOST spectral data to construct a dataset. This dataset has a time span of one month and contains unmerged red- and blue-arms data. 

As can be seen in Figure \ref{FIG:5}, the LAMOST observation range spans $5^{\circ}$, consists of 16 spectrographs. Figure \ref{FIG:6} illustrates the spatial distribution and classification of fibers within the central six spectrographs. It comprises 1,500 fibers, with the HD023750N350938V02 observational program yielding 1,147 red-arm spectra (including 129 sky spectra, 978 target fiber spectra, and 40 unusable spectra); while the blue-arm acquisition produced 1,147 spectra (134 sky spectra, 978 target fiber spectra, and 35 unusable fiber spectra). The composite dataset incorporating both spectra arms contains 2,219 distinct spectral observations. Similarly, datasets are constructed using four additional dark night observational programs: HD094551N184100B02, GAC066N10B1, GAC098N14B1, and GAC098N14B2, comprising a total of 11143 spectra. We also add several observational data on bright nights and present two testplanid as testplanid-1 20170210 and testplanid-2 20170211.

It should be noted that the data in this paper come from low-resolution observations conducted on the evening of January, 2017, and the data from the 31st are chosen for exhibition. Simultaneously, since the fitted sky background has been subtracted from the published data \citep{3}, this paper obtains the spectral data directly from the 2D CCD images (size 4136 × 4096) acquired from the observations to improve the estimation of sky background. A series of steps as follows: (1) Subtracting the Bias and CCD bottom fluxes. (2) Correcting the spectral data using the flat data to eliminate transmittance differences between fibers. (3) Tracing the spectral data for each fiber to obtain the position of that fiber in the CCD image. (4) Extract the 250 spectral data based on step (3) and calibrate the wavelength using a calibration lamp spectrum. Our processed data and publicly available LAMOST data is presented in Figure \ref{FIG:new} systematic flux offset between blue and red spectral arms is observable in our data, attributable to the absence of flux calibration procedure. SMI performs background estimation separately for the red and blue arms, so this does not affect the final results. More importantly, the above data contain more emission line features compared to LAMOST public data, particularly in the red arm. This is due to the fact that LAMOST public data subtract the sky background during pipeline processing based on a fitted “Super sky” , our reduction process preserves original sky spectra components, consequently enabling more accurate sky background characterization and subsequent scientific analysis.

The emission lines within the wavelength range of 6700\AA\ to 9180\AA\ display a high degree of density and intensity, in contrast to the relatively sparse distribution observed in the other wavelength ranges. This pattern reflects the general uneven distribution of the data. Consequently, we use an adaptive segmentation approach for the spectral data based on the distribution of emission lines. Specifically, segments within the sparse emission line region are larger, while segments within the dense emission line region are smaller. The segmentation is outlined as follows:
\begin{equation}
    length_{fin}= log(\frac{length}{counts})
\label{Eq:10}
\end{equation}
\noindent where length represents the length of the segment, counts for the length of the data within the number of emission line features, when the ratio of the two tends to stabilize the end, to get the length of the segment $length_{fin}$ can be acquired, the segment length is determined from the beginning of the execution of the process to get the length of the next segment.

\subsection{Result and Discussion}

The model utilizes post-segmentationspectral data for experiments and is evaluated in the following three ways: (1) Accuracy of the Whole Spectrum. (2) Accuracy of the Sky Emission Lines. (3) Visualization of “Shared” Sky Background Results. In the absence of standardized sky background spectra data as reference labels for comparative analysis, we propose that the sky fiber data represents the authentic sky background at the observation site and can serve as a reliable evaluation benchmark. The SMI estimations obtained from the sky fiber position are compared with the acquired sky fiber data for comprehensive analysis. Reduced variance signifies a closer match between the estimated sky background spectrum and the observed sky background spectrum within the sky fiber, indicating the effectiveness of our approach. Simultaneously, we display the “shared” sky representation, reflecting the model's estimation of the shared component.

\begin{figure*}[!ht]
    \centering
    \includegraphics[width=1.0\textwidth]{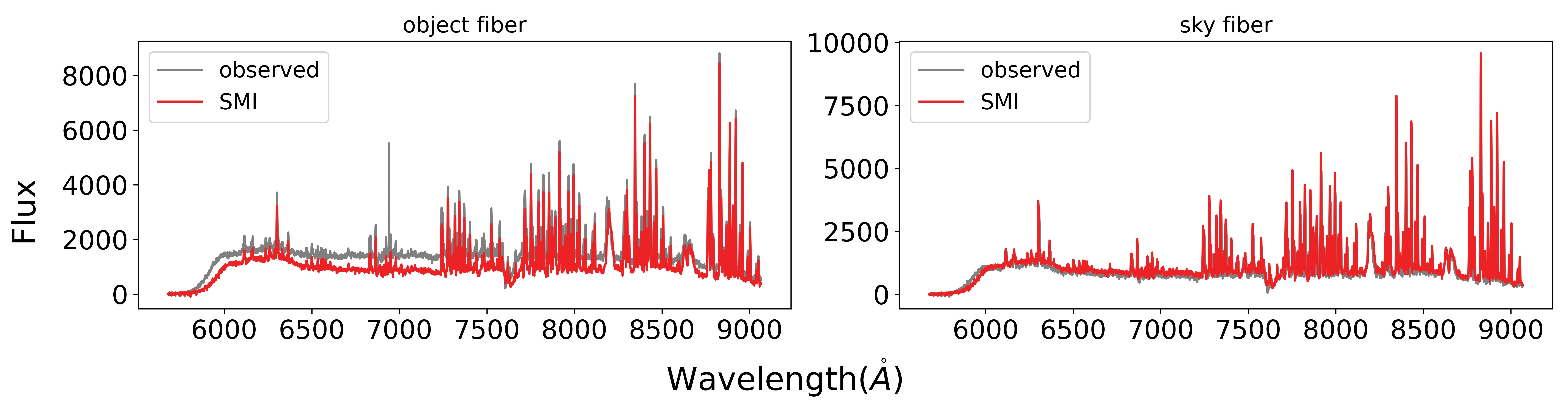}
    \caption{
     Derived Sky background. Results of SMI in target fiber and sky fiber, which in a same spectrometer. Where the gray data represents the observed data from the spectrometer and the red data represents the estimated results of the SMI.
    }    
    \label{FIG:diff}
\end{figure*}

\subsubsection{Accuracy of the Whole Spectrum}
\begin{figure*}
    \centering
    \includegraphics[width=1.0\textwidth]{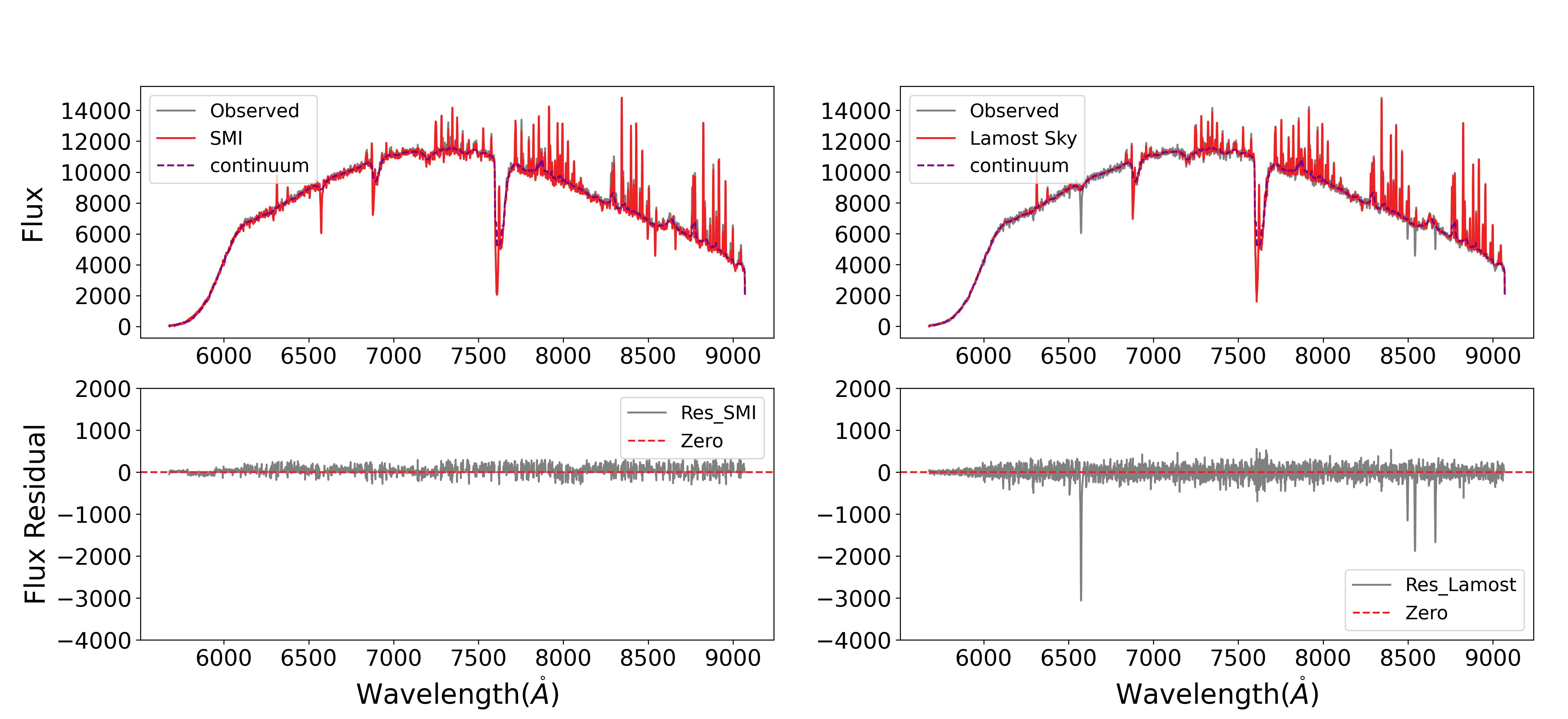}  
    \caption{
    Red-arm spectra, comparison of the sky background estimated by this method with the LAMOST “Super sky”.
    A comparison is made between the estimated sky background spectra and LAMOST sky spectra with the observed spectra. The x-axis denotes the wavelength while the y-axis represents the spectral flux.The observed sky spectra are depicted by a gray solid line, while the sky background obtained from the estimation based on mutual information and the sky data from LAMOST are represented by red solid lines.The bottom images display the residuals of the proposed method and the observed spectra.
    }    
    \label{FIG:7}
\end{figure*}

\begin{figure*}[!ht]
    \centering
    \includegraphics[width=1.0\textwidth]{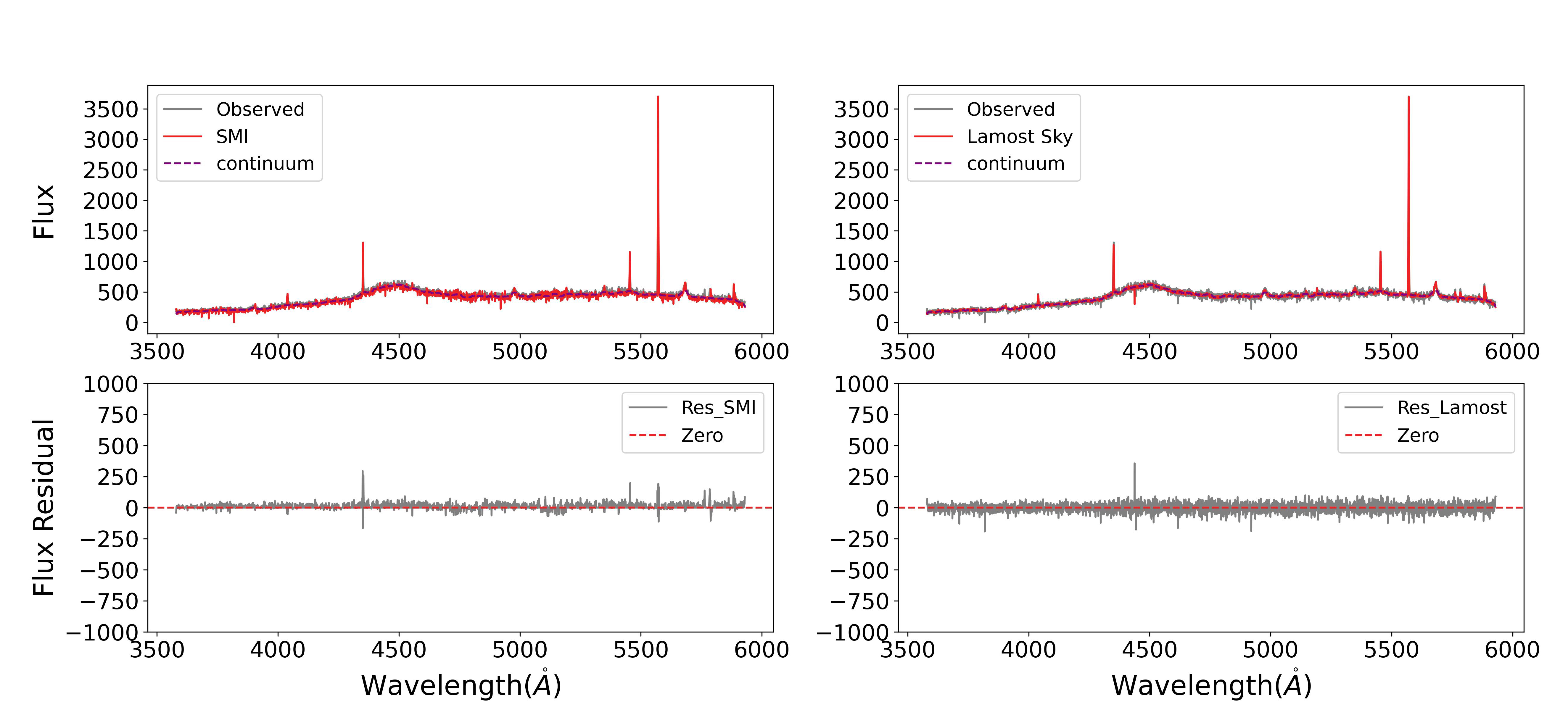}
    \caption{
    Blue-arm spectra, comparison of the sky background estimated by this method with the LAMOST “Super sky”.
    A comparison is made between the estimated sky background spectra and LAMOST sky spectra with the observed spectra. The x-axis denotes the wavelength while the y-axis represents the spectral flux. The observed sky spectra are depicted by a gray solid line, while the sky background obtained from the estimation based on mutual information and the sky data from LAMOST are represented by red solid lines.The bottom images display the residuals of the proposed method and the observed spectra.
    }    
    \label{FIG:8}
\end{figure*}

Figure \ref{FIG:diff} presents the sky background derived from SMI model on target and sky fibers. The target fiber result indicating significant discrepancies from observational data due to the inclusion of target object information, while the sky fiber estimations exhibit remarkable consistency with observed values. Given the current absence of standardized evaluation metrics for sky background assessment, this study employs a comparative analysis between the sky fiber results and LAMOST super sky spectra to validate the methodological efficacy. The residuals between the results of the mutual information-based method and the observed sky are calculated and compared, as depicted in Figures \ref{FIG:7} and \ref{FIG:8}. Compared with the "Super sky" of LAMOST, it can be observed that the sky background obtained by SMI on the sky fiber has a smaller residual from the actual observed sky data. This is particularly the case in wavelength with fewer emission lines, such as at the 4000\AA\ and 5000\AA\ in the blue end, and at the 6000\AA\ in the red end. In the bands with dense emission lines, the residuals of SMI are also lower than those of the LAMOST results.

\begin{figure*}[!ht]
        \centering
        \includegraphics[width=1.0\textwidth]{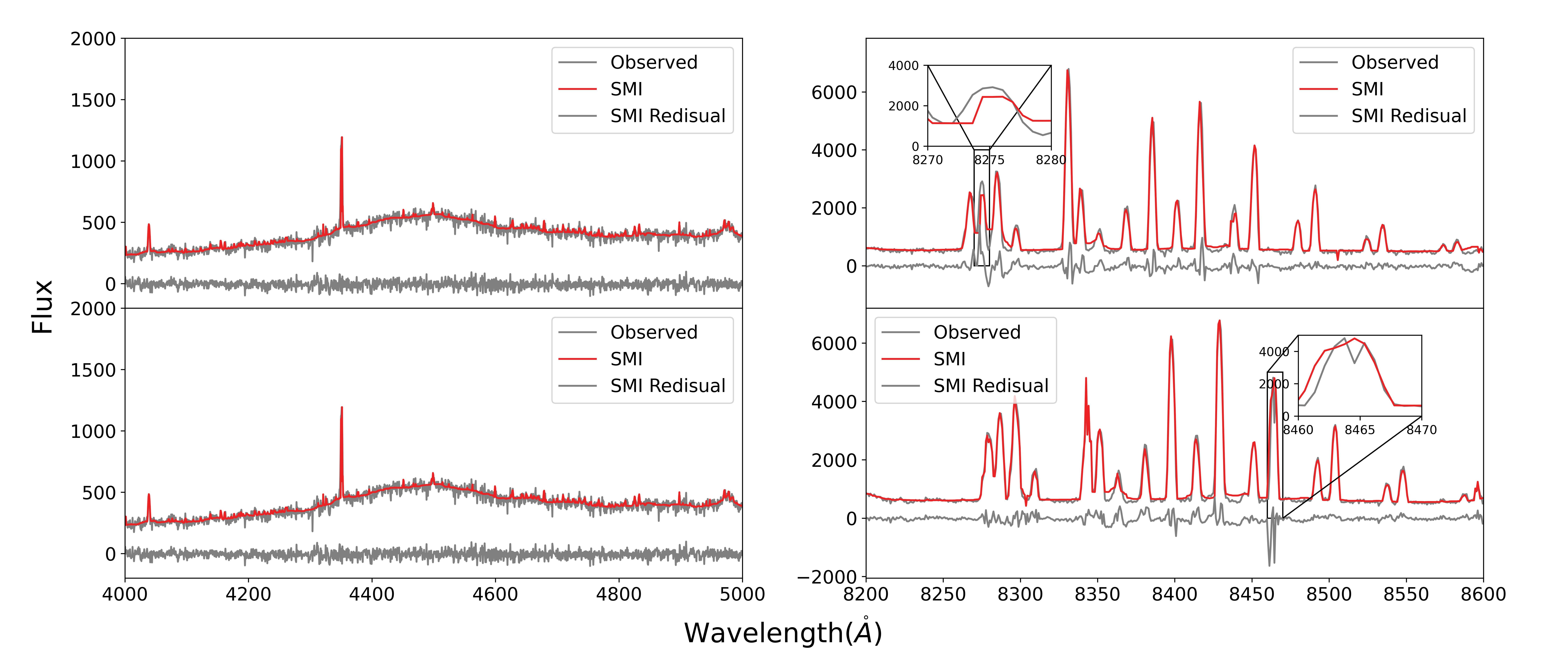}
        \caption{
        Constructed sky background in the band with emission lines. Data obtained from HD023750N350938V02 observation. x represents wavelength, while y for flux. The gray solid line represents observed sky data, while the red solid line represents constructed sky data. The residual between the two is also shown in the figure.
        }    
        \label{FIG:13}
\end{figure*}

\begin{figure}[!ht]
        \centering
        \includegraphics[width=1.0\columnwidth]{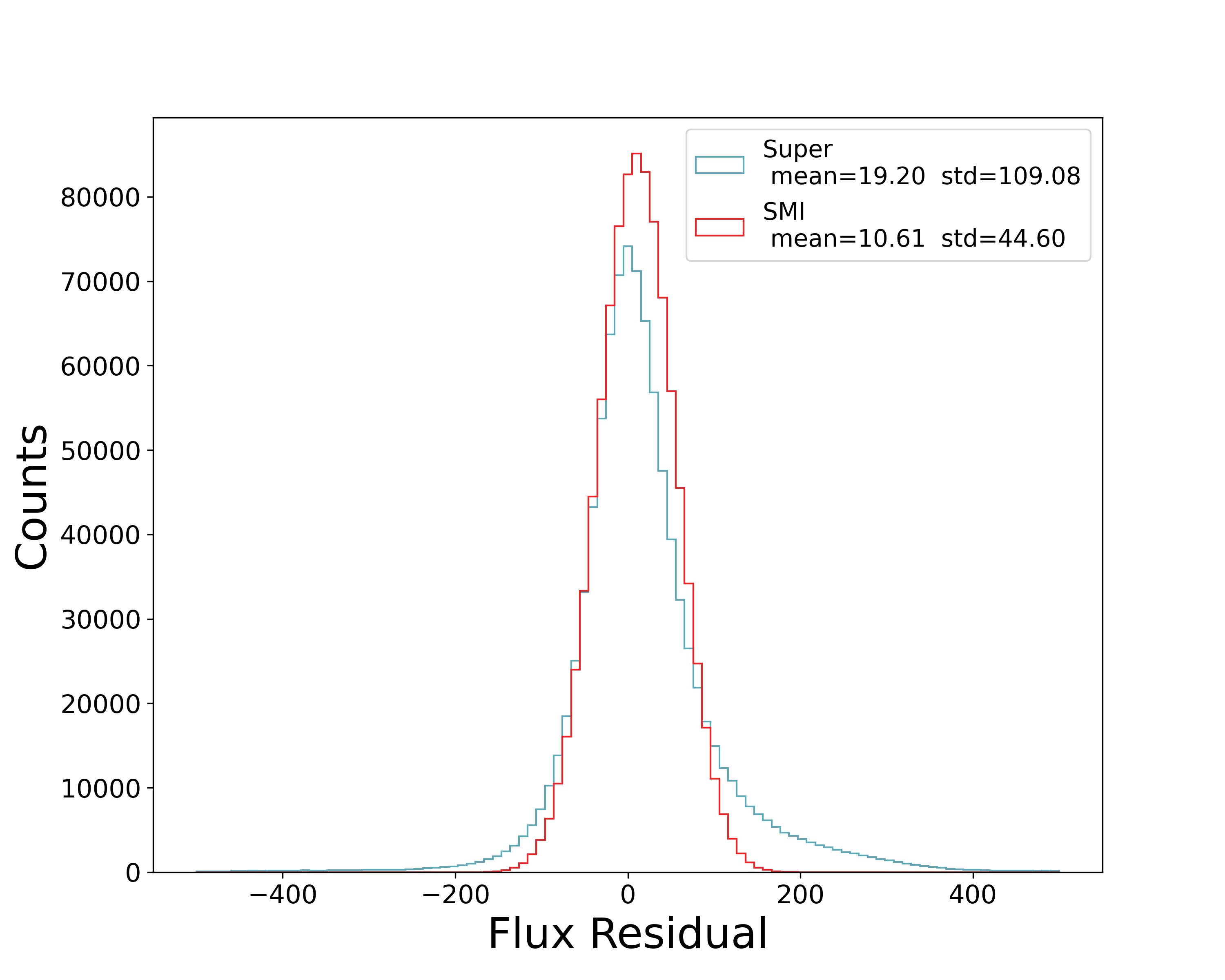}
        \caption{
        Histogram distribution of the six spectrometer construction datasets. The blue line represents the residual values of the LAMOST sky background, while the red line corresponds to the residuals derived from the sky background estimation method based on mutual information. The term "mean" in the upper right corner denotes the average value of the sky background residuals, while "std" indicates the standard deviation of these residuals.
        }    
        \label{FIG:9}
\end{figure}

\begin{table}
\begin{center}
\caption[]{ \centering Statistics of sky background spectra}\label{tab:1}
 \begin{tabular}{cccccc}
    \hline\noalign{\smallskip}
     & & \multicolumn{2}{c}{LAMOST} & \multicolumn{2}{c}{SMI}\\
    planid & spec & MAE & RMSE & MAE & RMSE\\
    \hline\noalign{\smallskip}
    \multirow{6}{*}{\rotatebox{90}{\shortstack{HD023750N3\\50938V02}}} & 03 & 31.73 & 76.75 & 15.25 & 36.82\\
    ~ & 04 & -0.61 & 72.99 & 0.21 & 44.70\\
    ~ & 05 & 6.44 & 108.48 & 4.41 & 71.22\\
    ~ & 08 & 36.09 & 102.71 & 20.14 & 73.65\\
    ~ & 09 & 21.01 & 78.86 & 10.44 & 34.33\\
    ~ & 15 & 17.61 & 71.40 & 13.10 & 50.34\\
    \hline\noalign{\smallskip}
    \multirow{6}{*}{\rotatebox{90}{\shortstack{HD094551N1\\84100B02}}} & 03 & 2.11 &	92.14 &	2.41 & 89.69\\
    ~ & 04 & -42.58	& 164.51 & -44.10 & 128.14\\
    ~ & 05 & 95.85 & 115.28 & 94.12 & 189.19\\
    ~ & 08 & -19.04 & 235.12 & -15.75 & 164.82\\
    ~ & 09 & 2.88 & 69.46 & 2.47 & 68.49\\
    ~ & 15 & 7.10 & 94.36 & 5.51 & 88.64\\
    \hline\noalign{\smallskip}
    \multirow{6}{*}{\rotatebox{90}{\shortstack{GAC066N1\\0B1}}} & 03 & 6.00&95.15&5.75&94.10\\
    ~ & 04 & -73.58	&235.82&-60.40&211.70\\
    ~ & 05 & 24.44&	135.97&	20.34&	128.77\\
    ~ & 08 & 22.05&	138.49&	20.19&	117.40\\
    ~ & 09 & -2.99&	73.84&	-1.79&	68.88\\
    ~ & 15 & 1.69&	172.89&	2.77&	168.49\\
    \hline\noalign{\smallskip}
    \multirow{6}{*}{\rotatebox{90}{\shortstack{GAC098N1\\4B1}}} & 03 & 6.82&	98.70&	6.57&	97.43\\
    ~ & 04 & 132.79&	250.85&	112.70&	216.24\\
    ~ & 05 & 115.57&	275.19&	107.73&	255.94\\
    ~ & 08 & 20.01&	186.68&	15.75&	150.84\\
    ~ & 09 & 15.29&	86.70&	10.93&	75.98\\
    ~ & 15 & 8.46&	84.86&	8.07&	84.54\\
    \hline\noalign{\smallskip}
    \multirow{6}{*}{\rotatebox{90}{\shortstack{GAC098N1\\4B2}}} & 03 & 5.14&	99.24&	5.06&	98.36\\
    ~ & 04 & 137.55&	171.33&	126.76&	110.21\\
    ~ & 05 & 154.16&	249.36&	154.50&	206.33\\
    ~ & 08 & 8.01&	116.63&	6.16&	103.85\\
    ~ & 09 & 14.41&	75.90&	8.49&	73.66\\
    ~ & 15 & 15.20&	181.36&	11.66&	149.43\\
    \hline\noalign{\smallskip}
    \multirow{6}{*}{\rotatebox{90}
    {\shortstack{testplanid-1\\20170210}}} 
      & 03 & 181.34 & 421.43 & 190.22 & 410.32\\
    ~ & 04 & -173.95 & 183.65 & -160.87 & 177.55\\
    ~ & 05 & -132.52 & 145.86 & -133.62 & 140.67\\
    ~ & 08 & 131.36 & 182.55 & 102.98 & 121.05\\
    ~ & 09 & -117.18 & 176.50 & -110.93 & 180.07\\
    ~ & 15 & 113.14 & 164.71 & 104.13 & 153.88\\
    \hline\noalign{\smallskip}
    \multirow{6}{*}{\rotatebox{90}
    {\shortstack{testplanid-2\\20170211}}} 
      & 03 & 187.29 & 290.29 & 155.47 & 243.88\\
    ~ & 04 & 264.77	& 435.33 & 270.93 & 400.82\\
    ~ & 05 & 265.99 & 339.85 & 159.08 & 230.43\\
    ~ & 08 & 119.44 & 308.99 & 109.34 & 252.81\\
    ~ & 09 & 184.61 & 144.88 & 153.22 & 120.93\\
    ~ & 15 & 108.39 & 194.69 & 84.33 & 105.72\\
    \hline\noalign{\smallskip}
\end{tabular}
\end{center}
\tablecomments{0.86\columnwidth}{We show the results from the center's six spectrometers in a single observation, as described in the Data Reduction section.}
\end{table}

We also exhibit the details for several spectral wavelength. As can be seen in left of Figure \ref{FIG:13}, this segment of spectral data has a few emission lines and a small residual, indicating that the estimation result is close to the observed data. This is because we estimate the emission lines and continuous spectra separately. The model is used to learn the information of emission lines with large variations, while the continuous spectrum with smaller variations is fitted by the neighborhood median method. Therefore, both the emission lines and continuous spectra in the results are close to the observed spectra. As shown in right of Figure \ref{FIG:13}, this segment of spectra has dense emission lines. Our result has a depression at the peak of the emission line, which leads to a relatively large residual. In addition, the width of some emission lines is relatively narrow. We analyze that this is because SMI sets a small kernel during training and simultaneously adds an attention mechanism to focus on the information at the emission line, making it more sensitive to it. And there are subtle differences in the sky background, which are mainly reflected in the position and flux of the emission lines. At the same time, during the observation process, due to the influence of various factors on the optical fiber, the spectra is prone to noise, which is particularly obvious at the emission lines. Therefore, the residual is relatively large at the dense emission line area.

The residual distributions of each spectrum are further counted in this paper, and the results are shown in Figure \ref{FIG:9}. It can be seen that the residuals obtained by SMI are more concentrated on zero and have smaller variances and means. It shows that the sky background obtained in the dataset better fits the observed data than the Super sky of LAMOST. This indicates that the results obtained by SMI are closer to the observed sky background. 

\begin{table}
\begin{center}
\caption[]{ \centering Statistics of sky background spectra at emission lines }\label{tab:2}
 \begin{tabular}{cccccc}
    \hline\noalign{\smallskip}
    & & \multicolumn{2}{c}{LAMOST} & \multicolumn{2}{c}{SMI}\\
    planid & spec & MAE & RMSE & MAE & RMSE\\
    \hline\noalign{\smallskip}
    \multirow{6}{*}{\rotatebox{90}{\shortstack{HD023750N3\\50938V02}}} & 03 & 31.60 & 77.26 & 20.99 & 54.50\\
    ~ & 04 & -4.19 & 79.44 & 2.45 & 50.80\\
    ~ & 05 & 6.12 & 108.56 & 4.94 & 70.39\\
    ~ & 08 & 34.73 & 104.258 & 17.56 & 78.81\\
    ~ & 09 & 21.50 & 79.63 & 10.88 & 36.64\\
    ~ & 15 & 22.55 & 144.19 & 17.47 & 104.52\\
    \hline\noalign{\smallskip}
    \multirow{6}{*}{\rotatebox{90}{\shortstack{HD094551N1\\84100B02}}} & 03 & 1.94&93.35&1.49&92.78\\
    ~ & 04 & -61.43&205.04&-59.41&187.75\\
    ~ & 05 & 109.25&	276.43&	104.74&	257.78\\
    ~ & 08 & -38.15&	376.43&	-24.90&	287.53\\
    ~ & 09 & 2.73&	69.71&	2.75&	66.44\\
    ~ & 15 & 6.43&	103.16&	5.37&	93.66\\
    \hline\noalign{\smallskip}
    \multirow{6}{*}{\rotatebox{90}{\shortstack{GAC066N1\\0B1}}} & 03 & 6.24&95.45&6.00&95.15\\
    ~ & 04 & -103.07&	291.85&	-100.06&	285.77\\
    ~ & 05 & 102.81&	400.04&	97.50&	387.85\\
    ~ & 08 & 24.85&	209.80&	22.17&	204.78\\
    ~ & 09 & -3.14&	73.92&	-2.15&	69.74\\
    ~ & 15 & 1.41&	196.09&	2.89&	187.98\\
    \hline\noalign{\smallskip}
    \multirow{6}{*}{\rotatebox{90}{\shortstack{GAC098N1\\4B1}}} & 03 & 6.39&98.76&6.07&98.70\\
    ~ & 04 & 206.05&	294.52&	195.76&	269.30\\
    ~ & 05 & 235.11&	325.35&	206.15&	294.74\\
    ~ & 08 & 87.54&	219.99&	79.36&	198.67\\
    ~ & 09 & 15.32&	86.71&	10.21&	75.97\\
    ~ & 15 & 8.09&	85.07&	7.96&	84.74\\
    \hline\noalign{\smallskip}
    \multirow{6}{*}{\rotatebox{90}{\shortstack{GAC098N1\\4B2}}} & 03 & 5.59&	103.72&	5.07&	98.70\\
    ~ & 04 & 134.01&	222.12&	106.16&	215.77\\
    ~ & 05 & 237.73&	251.46&	203.59&	222.16\\
    ~ & 08 & 11.53&	136.11&	10.01&	115.46\\
    ~ & 09 & 14.26&	75.92&	11.71&	64.46\\
    ~ & 15 & 46.85&	238.39&	44.58&	207.12\\
    \hline\noalign{\smallskip}
    \multirow{6}{*}{\rotatebox{90}
    {\shortstack{testplanid-1\\20170210}}} 
      & 03 & 96.32 & 170.54 & 108.34 & 121.32\\
    ~ & 04 & -137.30 & 195.89 & -173.52 & 145.47\\
    ~ & 05 & -127.89 & 150.43 & -117.14 & 176.04\\
    ~ & 08 & 233.47 & 254.13 & 113.71 & 126.67\\
    ~ & 09 & -294.79 & 277.04 & -258.61 & 272.03\\
    ~ & 15 & 178.67 & 214.67 & 177.67 & 189.71\\
    \hline\noalign{\smallskip}
    \multirow{6}{*}{\rotatebox{90}{\shortstack{testplanid-2\\20170211}}} 
      & 03 & 119.58 & 214.66 & 181.95 & 121.65\\
    ~ & 04 & 124.76 & 206.13 & 154.47 & 133.13\\
    ~ & 05 & 183.23 & 192.11 & 114.71 & 123.77\\
    ~ & 08 & 102.04 & 209.13 & 194.46 & 199.11\\
    ~ & 09 & 116.03 & 196.55 & 73.65 & 121.32\\
    ~ & 15 & 119.80 & 177.46 & 131.43 & 179.61\\
    \hline\noalign{\smallskip}
\end{tabular}
\end{center}
\tablecomments{0.86\columnwidth}{We show the results from the center's six spectrometers in a single observation, as described in the Data Reduction section.}
\end{table}

Then we calculate the mean absolute error (MAE) and the root mean squared error (RMSE) between the two model results and the observed spectra to assess the goodness of fitting of the model results. Table \ref{tab:1} presents the residuals of spectral data from six spectrometers. Taking the HD023700N350938V02 observation as an example, the MAE values are pronounced between the SMI results and sky fiber data for 03, 08, and 09. Concurrently, the RMSE decreased by nearly half, indicating that the SMI results indicating closer agreement with observational data and enhanced stability on these three spectrographs. For 04, 05, and 15, MAE and RMSE from both SMI and LAMOST nearly zero, suggesting satisfactory sky background estimation in both methods.The results of the final two observation programs results indicating differences between the SMI and LAMOST. Compared to LAMOST pipeline, the results some spectrographs, particularly the 04 on testplanid-2, exhibit inferior performance. These variations might attribute to incomplete extraction of mutual information from the sky background, or biases introduced by the sky fiber position. Nevertheless, the SMI exhibits smaller RMSE values, demonstrating superior stability and fewer outliers. It can be observed that the residuals of the proposed method are closer to zero to the actual observed sky spectra, and smaller statistical results reflects better estimation: a smaller MAE implies the closer proximity to observations, while a smaller RMSE indicates closer proximity to the observed spectra. 

\subsubsection{Accuracy of the Sky Emission Lines}

\begin{figure}
    \centering
    \includegraphics[width=1.0\columnwidth]{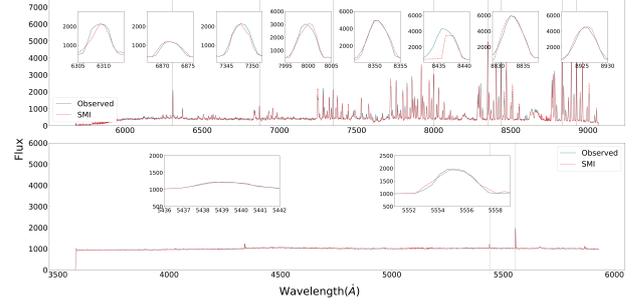}
    \caption{
    Sky emission lines at the red and blue arms. The figure has identified the specific positions of ten sky emission lines corresponding to Hg, O, and OH. The observed spectrum is represented by the blue line, while the sky background derived from mutual information are depicted by the orange dashed line. Each of the ten emission line positions is clearly labeled, with five positions selected for closer examination through zoom analysis.
    }    
    \label{FIG:10}
\end{figure}
\begin{figure}
    \centering
    \includegraphics[width=1.0\columnwidth]{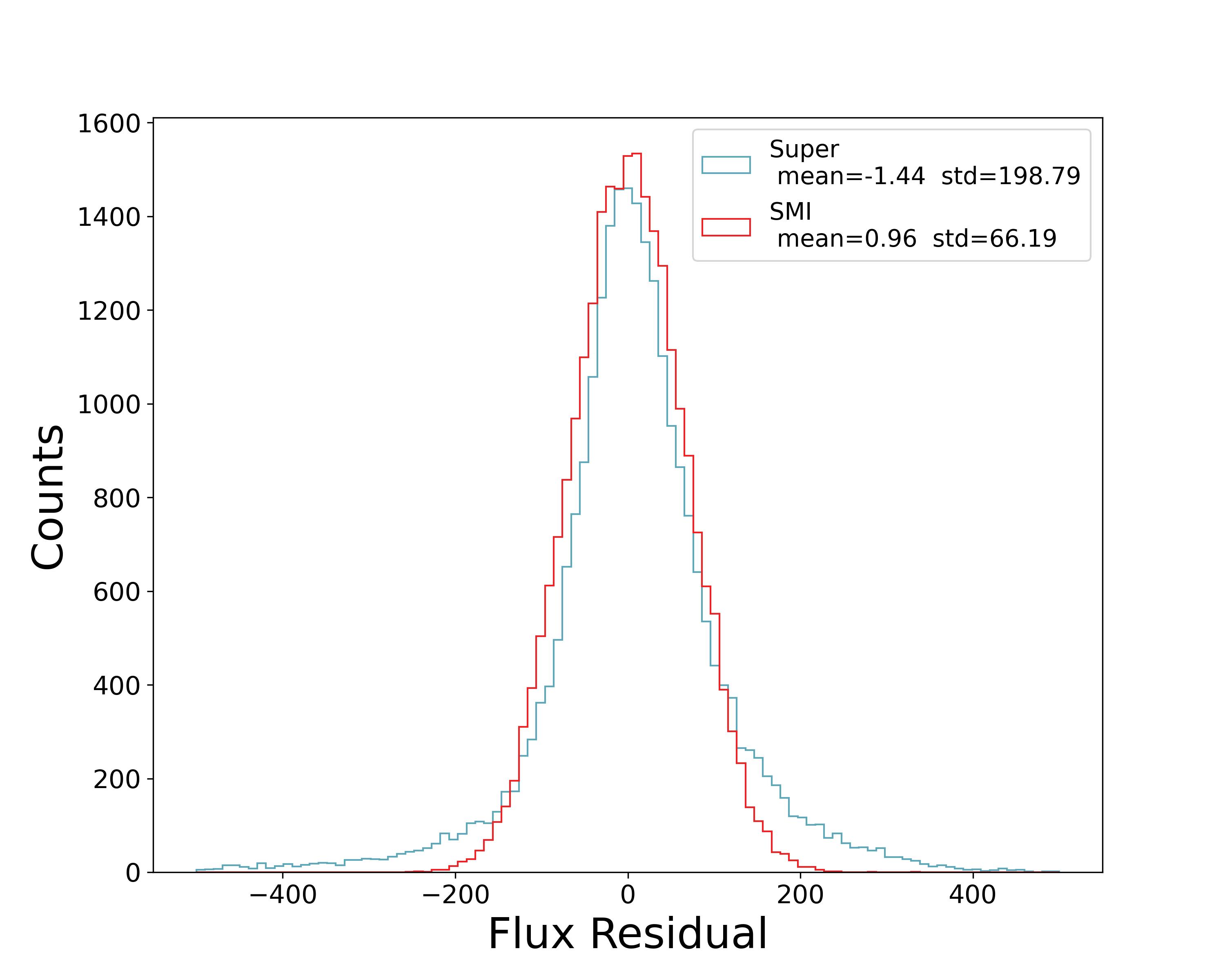}
    \caption{
    Histogram distribution of sky background emission lines. We use 3$\sigma$ method to obatin the emission line. The LAMOST sky background residuals are represented by the blue line, while the residuals of the sky background estimation method based on mutual information are depicted in red. The average and dispersion of the residual outcomes for both techniques are indicated in the top right corner.
    }    
    \label{FIG:11}
\end{figure}

Given that the sky background is mainly characterized by the emission lines, a more detailed presentation of the sky background results at these emission lines is carried out. We choose the 3$\sigma$ detection method to obtain the emission lines in the spectra instead of calculating the residuals based on the celestial body emission lines. As can be seen in Figure \ref{FIG:10}, the flux and location for most of the emission lines are well reproduced by our results, but there are still a few regions where the predicted emission line profiles deviate from the observed ones. As illustrate in Figure \ref{FIG:11}, SMI exhibits smaller variances and means, and is more concentrated around zero values, indicating that SMI fits the actual observed emission line better. 

\begin{table}
\begin{center}
\caption[]{ \centering Statistics of sky background spectra at several emission lines }\label{tab:3}
 \begin{tabular}{ccccc}
    \hline\noalign{\smallskip}
    & \multicolumn{2}{c}{LAMOST} & \multicolumn{2}{c}{SMI}\\
    Index of pixel & MAE & RMSE & MAE & RMSE\\
    \hline\noalign{\smallskip}
    6506($\pm7$) & -0.66 & 73.32 & 1.05 & 60.42\\
    7032($\pm7$) & -0.30 & 73.03 & 0.54 & 62.87\\
    7245($\pm7$) & -0.28 & 73.24 & 0.67 & 70.51\\
    7438($\pm7$) & -0.45 & 73.23 & 0.21 & 57.89\\
    8377($\pm7$) & -0.72 & 73.11 & 0.70 & 59.14\\
    \hline\noalign{\smallskip}
\end{tabular}
\end{center}
\end{table}

The MAE and RMSE of SMI and Super sky on emission lines of six spectrographs are shown in Table \ref{tab:2}. It can be seen that the residuals of the proposed method have smaller statistical results, which reflects better estimation: not only closer proximity to the observed spectra overall, but also to the presence of fewer and smaller outliers. At the same time, we note that the residuals at the emission lines are larger than on the full spectrum for both Super Sky and SMI, and we suspect that this is for similar reasons as in Figure \ref{FIG:13}. Subsequently, in order to show the results more detail, we calculated the emission line residuals at five positions have high intensity, instead of at all emission lines across the spectrum. We calculated for 14 pixel widths in the vicinity of the emission lines. The results are shown in Table \ref{tab:3}, it can be observed from the table that Super Sky yields good results, with the residuals of the emission lines being close to zero at these points. Additionally, the SMI results also have smaller residuals and exhibit less variance.

\subsubsection{Visualization of “Shared” Sky Background Results}

\begin{figure*}[!ht]
        \centering
        \includegraphics[width=1.0\textwidth]{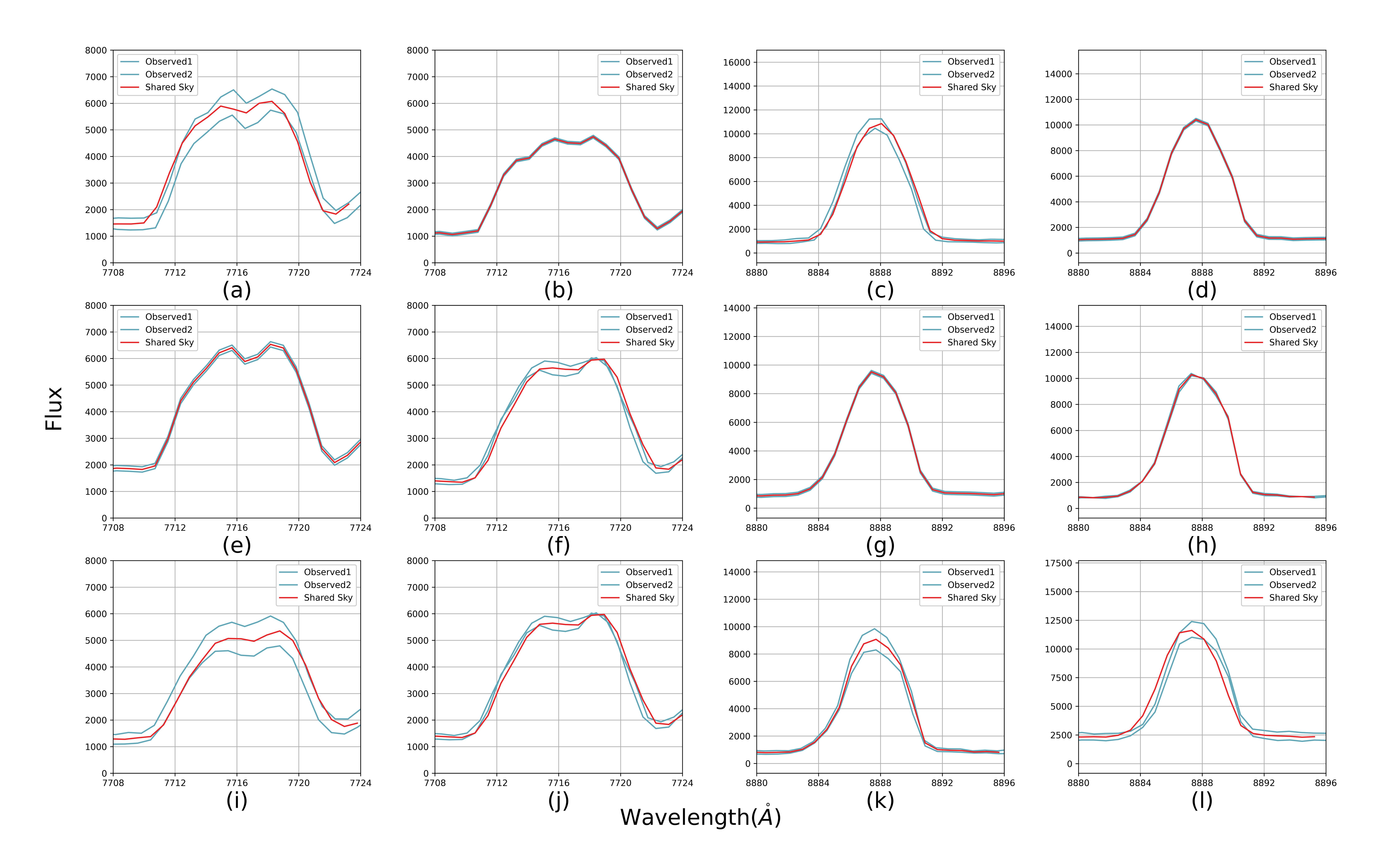}
        \caption{
        “Shared" sky representation. The sky background has differences in emission line positions and intensities at different location, two distinct spectra display discrepancies in emission line intensities within the same wavelength range, as indicated by the blue solid lines. The red solid line represents the common sky representation based on the former two observed spectra data.
        }    
        \label{FIG:12}
\end{figure*}

This section presents the results of the "shared" sky representation. As illustrated in Figure \ref{FIG:12}, the adjacent fibers No.9 and 17 in GAC071N47B1 are selected to indicating the "Shared” sky background results. Given that one set of spectra contains more emission lines, this section concentrates on a specific emission line for analysis. The solid blue line in the figure shows the observed spectra, where the characteristics of the emission lines in the same wavelength region are inconsistent, exhibiting minor variations in both position and flux. The solid red line in the figure shows a representation of the sky shared components obtained in the first stage of the model.

We use the shared representation to capture the common sky information in multiple data, so the results should include the relevant features of the sky emission lines in the two data, containing the position and flux information. As presented in (b), (d), (e), (g) and (h), the spatial alignment of emission features in “shared” sky precisely corresponds with their counterparts in the blue reference data, indicated by red and blue spectral profile overlaps, confirming effective acquisition of sky emission line parameters. This spatial congruence, coupled with consistent width , indicates successful retention of spatial details for dual emission lines. While (a), (c), (i), and (k) show maintained positional accuracy between "shared" sky and blue data, flux discrepancies persist in the spectra data. In (j),  flux depression at emission line peaks. This anomaly attributable to significant flux gradient variations, could be mitigated through post-processing smoothing techniques. Notably, the "shared" representation indicating robust capability in extracting common spectra features across multiple observations while preserving line parameters. The analysis confirms the SMI effectiveness in maintaining spatial consistency for emission line positions and profile widths. Overall, SMI effectively captures the common information within several spectral data.

\section{Summary} \label{sect 4}

This paper proposed a model SMI employ mutual information by incremental training approach. Distinct from the current sky subtraction methods, it employs mutual information to learn from the spectra of all the fibers on the same spectrometer for estimating the sky background. Thus, it can fully utilize the sky information contained in the spectra for sky background estimation and obtain better results. We utilized LAMOST spectral data to construct the dataset for our experiments. The results show that the sky background obtained by SMI is closer to the observed sky spectra with fewer outliers. Meanwhile, the "shared" representations effectively capture the common information within several spectra.

In the present research, our main focus has been on comparing the results with the actual observed sky background. However, we have not yet performed downstream tasks, such as performing parameter measurements of target stars following sky background subtraction. Furthermore, there is a pattern of distribution of the sky background not only in space but also in the time domain, like \citep{32} analyzed the temporal variations of the sky background using time-series data and found that the sky background exhibits certain regularities on time scales. Future research could build on this study and fill this gap by further exploring the influences of the sky background and its impact on subsequent measurement tasks. 

The SMI proposed in this paper effectively utilizes spectra of all the fibers from a single observation and integrates a deep learning network to explore the relationship between sky backgrounds. \citet{9} proposed that sky component in the target spectrum is established to comprise two distinct constituents: continuum and emission line features; \citet{32} identified discernible temporal patterns in the sky background through cluster analysis; \citet{37} highlighted the necessity of establishing standardized protocols for sky subtraction accuracy; \citet{17} highlighted gradient variation patterns in the sky background. These seminal investigations collectively laid the theoretical groundwork for subsequent sky background studies and proposed an operational framework for background elimination. The innovative use of mutual information for sky background subtraction is expected to provide an important foundation for future research in similar fields and promote further research in this area.

\begin{acknowledgements}
The work is supported by 
the National Natural Science Foundation of China (Grant Nos. 12473106, 12473105), 
Projects of Science and Technology Cooperation and Exchange of Shanxi Province (Grant Nos. 202204041101037, 202204041101033), 
the central government guides local funds for science and technology development (YDZJSX2024D049),
and the science research grant from the China Manned Space Project with No. CMS-CSST-2021-B03.

We thank the Guo Shou Jing Telescope (the Large Sky Area Multi-Object Fiber Spectroscopic Telescope, LAMOST).
LAMOST is a National Major Scientific Project built by the Chinese Academy of Sciences. Funding for the project has been provided by the National Development and Reform Commission.
\end{acknowledgements}
  
\bibliographystyle{raa}
\bibliography{main}

\end{document}